\documentclass[draft]{agujournal2019}
\usepackage{url} %this package should fix any errors with URLs in refs.
\usepackage[inline]{trackchanges} %for better track changes. finalnew option will compile document with changes incorporated.
\usepackage{soul}
\usepackage{amsmath}
\usepackage{todonotes}

\draftfalse

\journalname{Water Resources Research}

\begin{document}

%%%%%%%%%%%%%%%%%%%%%%%%%%%%%%%%%%%%%%%%%%%%%%%
%  TITLE
%
% (A title should be specific, informative, and brief. Use
% abbreviations only if they are defined in the abstract. Titles that
% start with general keywords then specific terms are optimized in
% searches)
%
%%%%%%%%%%%%%%%%%%%%%%%%%%%%%%%%%%%%%%%%%%%%%%%

% Example: \title{This is a test title}

\title{An End-to-End Differentiable, Graph Neural Network–Embedded Pore Network Model for Permeability Prediction}

%%%%%%%%%%%%%%%%%%%%%%%%%%%%%%%%%%%%%%%%%%%%%%%
%
%  AUTHORS AND AFFILIATIONS
%
%%%%%%%%%%%%%%%%%%%%%%%%%%%%%%%%%%%%%%%%%%%%%%%

% Authors are individuals who have significantly contributed to the
% research and preparation of the article. Group authors are allowed, if
% each author in the group is separately identified in an appendix.)

% List authors by first name or initial followed by last name and
% separated by commas. Use \affil{} to number affiliations, and
% \thanks{} for author notes.
% Additional author notes should be indicated with \thanks{} (for
% example, for current addresses).

% Example: \authors{A. B. Author\affil{1}\thanks{Current address, Antartica}, B. C. Author\affil{2,3}, and D. E.
% Author\affil{3,4}\thanks{Also funded by Monsanto.}}

\authors{Qingqi Zhao\affil{1}, Heng Xiao\affil{1, 2}}

% \affiliation{1}{First Affiliation}
% \affiliation{2}{Second Affiliation}
% \affiliation{3}{Third Affiliation}
% \affiliation{4}{Fourth Affiliation}

\affiliation{1}{Stuttgart Center for Simulation Science, University of Stuttgart, Stuttgart, 70569, Germany}

\affiliation{2}{Institut für Thermodynamik der Luft- und Raumfahrt (ITLR), University of Stuttgart, Stuttgart, 70569, Germany}
%(repeat as many times as is necessary)

% Corresponding author mailing address and e-mail address:

% (include name and email addresses of the corresponding author.  More
% than one corresponding author is allowed in this LaTeX file and for
% publication; but only one corresponding author is allowed in our
% editorial system.)

% Example: \correspondingauthor{First and Last Name}{email@address.edu}

\correspondingauthor{Heng Xiao}{heng.xiao@simtech.uni-stuttgart.de}

%%%%%%%%%%%%%%%%%%%%%%%%%%%%%%%%%%%%%%%%%%%%%%%
% KEY POINTS
%%%%%%%%%%%%%%%%%%%%%%%%%%%%%%%%%%%%%%%%%%%%%%%
%  List up to three key points (at least one is required)
%  Key Points summarize the main points and conclusions of the article
%  Each must be 140 characters or fewer with no special characters or punctuation and must be complete sentences

% Example:
% \begin{keypoints}
% \item	List up to three key points (at least one is required)
% \item	Key Points summarize the main points and conclusions of the article
% \item	Each must be 140 characters or fewer with no special characters or punctuation and must be complete sentences
% \end{keypoints}

\begin{keypoints}
\item Graph neural network learns pore‑scale conductance; its outputs flow into a pore network solver for bulk permeability estimation.
\item A discrete adjoint enables end-to-end training from bulk permeability data alone, eliminating the need for pore-scale labels.
\item Gradient analysis shows physically consistent sensitivities of pore scale features to bulk permeability, enhancing model interpretability.
\end{keypoints}

%%%%%%%%%%%%%%%%%%%%%%%%%%%%%%%%%%%%%%%%%%%%%%%
%
%  ABSTRACT and PLAIN LANGUAGE SUMMARY
%
% A good Abstract will begin with a short description of the problem
% being addressed, briefly describe the new data or analyses, then
% briefly states the main conclusion(s) and how they are supported and
% uncertainties.

% The Plain Language Summary should be written for a broad audience,
% including journalists and the science-interested public, that will not have 
% a background in your field.
%
% A Plain Language Summary is required in GRL, JGR: Planets, JGR: Biogeosciences,
% JGR: Oceans, G-Cubed, Reviews of Geophysics, and JAMES.
% see http://sharingscience.agu.org/creating-plain-language-summary/)
%
%%%%%%%%%%%%%%%%%%%%%%%%%%%%%%%%%%%%%%%%%%%%%%%

%% \begin{abstract} starts the second page

\begin{abstract}
Accurate prediction of permeability in porous media is essential for modeling subsurface flow. While pure data-driven models offer computational efficiency, they often lack generalization across scales and do not incorporate explicit physical constraints. Pore network models (PNMs), on the other hand, are physics-based and efficient but rely on idealized geometric assumptions to estimate pore-scale hydraulic conductance, limiting their accuracy in complex structures. To overcome these limitations, we present an end-to-end differentiable hybrid framework that embeds a graph neural network (GNN) into a PNM. In this framework, the analytical formulas used for conductance calculations are replaced by GNN-based predictions derived from pore and throat features. The predicted conductances are then passed to the PNM solver for permeability computation. In this way, the model avoids the idealized geometric assumptions of PNM while preserving the physics-based flow calculations. The GNN is trained without requiring labeled conductance data, which can number in the thousands per pore network; instead, it learns conductance values by using a single scalar permeability as the training target. This is made possible by backpropagating gradients through both the GNN (via automatic differentiation) and the PNM solver (via a discrete adjoint method), enabling fully coupled, end-to-end training. The resulting model achieves high accuracy and generalizes well across different scales, outperforming both pure data-driven and traditional PNM approaches. Gradient-based sensitivity analysis further reveals physically consistent feature influences, enhancing model interpretability. This approach offers a scalable and physically informed framework for permeability prediction in complex porous media, reducing model uncertainty and improving accuracy.
\end{abstract}

\section*{Plain Language Summary}
Predicting how easily fluids move through porous materials is important for understanding groundwater flow, oil recovery, and environmental cleanup. One way to do this is by using pore network models, which simulate fluid flow through a simplified network of pores and channels. While pore network models are based on physical principles and run efficiently, they often use oversimplified formulas to calculate how easily fluid can flow through individual pores, which limits their accuracy in complex geometry. In this study, we improve permeability prediction by combining physics-based modeling with machine learning. Specifically, we use a graph neural network to replace the simplified formulas in pore network models with data-driven predictions, based on detailed pore and throat geometry. The rest of the pore network solver remains unchanged, so the model still relies on known physical laws. The graph neural network is trained without needing detailed measurements at every pore and only uses a single overall permeability value for training. By optimizing the model end-to-end using gradients from both the neural network and the flow solver, we achieve accurate and generalizable predictions. This hybrid approach improves both accuracy and interpretability, and provides a scalable solution for modeling flow in complex porous media.

%%%%%%%%%%%%%%%%%%%%%%%%%%%%%%%%%%%%%%%%%%%%%%%
%
%  BODY TEXT
%
%%%%%%%%%%%%%%%%%%%%%%%%%%%%%%%%%%%%%%%%%%%%%%%

%%% Suggested section heads:
\section{Introduction}
%
% The main text should start with an introduction. Except for short
% manuscripts (such as comments and replies), the text should be divided
% into sections, each with its own heading.

% Headings should be sentence fragments and do not begin with a
% lowercase letter or number. Examples of good headings are:
Understanding subsurface flow behavior is critical for many applications, including geological carbon storage \cite{bickle2009geological}, underground hydrogen storage \cite{wang2024pore}, oil and gas extraction \cite{sakhaee2012gas}, and contaminant fate and transport in groundwater \cite{smith2024integrated}. Permeability is a fundamental property controlling fluid flow in porous media and is typically quantified using Darcy's law, which describes the ease with which fluids move through porous media~\cite{adeyemi2022determining}. Several numerical methods have been developed to simulate permeability. Among these methods, direct numerical simulation (DNS), such as the lattice Boltzmann method (LBM), provides the most accurate estimation of permeability \cite{mcclure2014novel,guo2022role}. DNS explicitly resolves the velocity field within the porous structure, allowing precise calculation of permeability based on Darcy's law. However, DNS is computationally intensive, demanding significant computational resources, memory, and time. 

% An alternative numerical approach is the pore network model (PNM), which offers a computationally efficient way to estimate permeability compared to DNS \cite{joekar2012analysis,raoof2010new,blunt2001flow}. PNM simplifies the porous medium by segmenting it into distinct pore regions, which are represented as nodes. These nodes are interconnected by links that denote the connectivity between pore pairs. Each node is characterized by specific features, such as pore diameter and pore length, estimated based on simplified assumptions of pore geometry shapes. Darcy's law is then applied to calculate the flow rates between each connected pore pair, enabling permeability estimation based on a coarsened flow field. Although PNM significantly reduces computational resources and speeds up the simulation, the accuracy of permeability predictions decreases due to the simplifications involved—particularly the assumption of simplified pore geometry shapes and the coarsening of the flow field.

Over the past decade, machine learning and deep learning have emerged as promising tools across many scientific and engineering applications, including aerospace \cite{duraisamy2019turbulence}, material science \cite{jones2021improved}, medical imaging \cite{erickson2017machine}, and geosciences \cite{wang2023application}. These techniques are particularly effective at capturing complex functional relationships between input and output data, obtained either through experiments or numerical simulations. The simplest approach for predicting permeability from porous media images involves extracting microstructural characteristics and using them as input to a multilayer perceptron (MLP) \cite{tian2021permeability,fu2023data,tembely2021machine,kamrava2020linking}. However, these features are typically computed at the whole-domain level, which provides only high-level statistical summaries of the porous medium. As a result, they may not adequately capture the local heterogeneity or the detailed geometric structure of the pore space, leading to limited predictive accuracy.

Deep learning, specifically convolutional neural networks (CNNs), has also been employed extensively to predict permeability from micro-CT images, such as digital rock images \cite{da2021deep,yang2024data}. CNNs excel in pattern recognition and image segmentation, making them particularly suitable for directly processing porous media images to predict permeability \cite{kang2024hybrid,liu2023uncertainty,caglar2022deep,tian2020surrogate}. One strategy to enhance CNN-based predictions is to incorporate additional physical information alongside the input images. \citeA{wu2018seeing} demonstrated improved permeability predictions by combining a CNN with physical parameters such as porosity and specific surface area in two-dimensional synthetic porous media images. Subsequent studies have extended this approach to more complex scenarios, including three-dimensional synthetic porous media and real sandstone samples \cite{tang2022predicting,garttner2023estimating,tembely2020deep}. Additionally, various innovative CNN architectures have been proposed to further improve permeability predictions \cite{zhang2022permeability,dos2023permeability}. For instance, \citeA{elmorsy2022generalizable} developed a CNN featuring an inception module with parallel paths utilizing different kernel sizes, enhancing the model's generalizability by capturing multiscale structural information. More advanced designs include combining CNNs with self-attention mechanisms and integrating additional physical information, achieving further improvements \cite{meng2023transformer}. Another distinct approach involves predicting the complete velocity field within the pore spaces first using CNNs and subsequently calculating permeability based on the predicted flow fields \cite{wang2021ml,santos2020poreflow, zhou2022neural}. Such method focuses more on flow field modeling rather than direct permeability estimation. One key limitation of CNN-based methods for permeability prediction from images is their high memory consumption, especially for large three-dimensional image domains. Recent studies have addressed this issue through various approaches, including point-cloud representations and multiscale techniques \cite{nabipour2024computationally,jiang2023upscaling}. \citeA{kashefi2021point} applied a point-cloud deep learning method, significantly reducing input size by focusing solely on boundary points between the solid matrix and pore spaces. Another innovative approach proposed by \citeA{elmorsy2023rapid} involves partitioning large images into smaller subsections, predicting the permeability of each subdomain using a lightweight CNN, and then using a multilayer perceptron to estimate the whole-domain permeability from the subdomain results. These approaches highlight practical methods for overcoming memory constraints in permeability predictions from large-scale porous media images.

Another method to improve computational efficiency in permeability prediction is the use of graph neural networks (GNNs)—a type of neural network specifically designed for graph-structured data \cite{wu2020comprehensive,scarselli2008graph,corso2024graph}. GNNs were developed to overcome the limitations of traditional CNNs in handling irregular structures and variable input sizes. They are well-suited for learning complex relationships between graph nodes and have been widely applied in tasks such as graph classification \cite{han2024topology} and link prediction \cite{zhang2018link}. Graph neural networks have been increasingly used to model and predict flow behavior in porous media~\cite{alzahrani2023pore,zhao2024rtg, KIM2025107497}. \citeA{zhao2025computationally} developed a GNN model that predicts porous media permeability directly from the graph. In this graph, pores are represented as nodes and throats as edges. The GNN then predicts permeability by processing the input graphs through node and edge features. One key advantage of this method is its flexibility. Since GNNs operate on node and edge features rather than fixed-size input grids, they can handle graphs of arbitrary size. Moreover, converting image data into pore level graphs significantly reduces memory usage, as clusters of pixels are compressed into single nodes with compact feature vectors.

% Similar to how the RANS solver uses Reynolds stresses as closure model parameters, 

Building on the same underlying graph-based representation, the pore network model (PNM) offers a physically motivated and computationally efficient framework for predicting permeability \cite{joekar2012analysis,raoof2010new,blunt2001flow}. It uses the hydraulic conductance between connected pore pairs as model parameters to predict permeability. Hydraulic conductance quantifies how easily fluid flows through each pore-throat connection. Given a pore network structure, permeability can be calculated by assigning hydraulic conductance values to all pore pairs and solving the pore network equations. However, hydraulic conductance is typically computed using simplified analytical formulas that assume idealized pore geometries. These geometric assumptions often fail to capture the true complexity of pore shapes, introducing uncertainty into the predictions. Several studies have attempted to improve the accuracy of hydraulic conductance estimation. For example, \citeA{liu2022pore} proposed an improved pore-throat segmentation algorithm based on local hydraulic resistance equivalence to enhance permeability prediction. Rather than refining the analytical formulas themselves, another line of work replaces them entirely with neural network models. \citeA{misaghian2022prediction}, for instance, used a CNN to predict the diffusional conductance, a parameter analogous to hydraulic conductance, from local pore images. They then used the predicted conductance values within a pore network model to compute the bulk-level formation factor. However, this approach requires training data obtained from high-fidelity simulations, such as lattice Boltzmann methods, for each individual pore pair. A single pore network can contain thousands of such pore pairs, each requiring its own simulation. As a result, preparing the training data becomes computationally expensive even for just one porous media sample. Moreover, unlike bulk permeability, pore-scale hydraulic conductance is impractical to measure experimentally, as individual pore pairs cannot be isolated and tested independently.

The reliance on pore-level training data presents a major challenge for deep learning applications in pore network modeling. To address this, we draw inspiration from recent advances in data-driven Reynolds-averaged Navier–Stokes (RANS) modeling. Rather than employing fully data-driven models, these approaches adopt hybrid physics–neural network frameworks, where neural networks are used to represent closure terms, such as Reynolds stresses or eddy viscosity. These learned components are then embedded within a RANS solver, forming part of a coupled partial differential equation (PDE) system \cite{strofer2021end,zhang2022ensemble}. In this framework, the model is trained not on the closure terms themselves, but by comparing the final outputs (velocity or pressure fields) of the RANS solver to reference data. These outputs are referred to as indirect data, because they are not directly predicted by the neural network model, but instead result from solving the physics-based equations using the neural network model’s output. In contrast, direct data refer to training data that correspond directly to the outputs of the neural network model, such as labeled Reynolds stress fields. While direct data simplifies model design and training, it is often difficult to obtain in practice, especially for complex or high-Reynolds-number flows. As a result, hybrid approaches that use neural networks to produce numerical model components, which are then passed into traditional solvers to match against indirect observations, offer a more practical and physically grounded alternative for many real-world applications. 

Following this concept, PNM can also be interpreted in terms of having direct and indirect data. Pore-scale hydraulic conductance, while representing high-fidelity information, is impractical to obtain experimentally and therefore constitutes direct data. In contrast, bulk permeability, which is readily measurable and more feasible to acquire across a diverse dataset of representative pore structures, can be treated as indirect data. An overview of the analogy between Reynolds-Avearged Navier--Stokes models and Pore Network Models in terms of direct and indirect observations is provided in Table~\ref{tab:rans-pnm-analogy}. In this work, we adopt a similar strategy to avoid relying on direct pore-scale training data. Inspired by the works in data-driven turbulence modeling~\cite{zhang2022ensemble}, we propose a graph neural network-embedded pore network model, where the graph neural network is used to infer the pore-scale hydraulic conductance needed to run the pore network model. Analogous to how neural networks are used in RANS modeling to represent closure terms (i.e., the Reynolds stresses), the graphical neural network in our framework predicts hydraulic conductance as the closure term in the pore network model. The goal in both RANS and PNM settings is to improve the neural network’s prediction of the closure term in a physical model. This introduces the challenge of optimizing the neural network not based on its outputs directly, but based on how those outputs influence the solution of the governing equations. While standard learning techniques can handle direct supervision, additional strategies are needed to account for the indirect supervision. One widely used solution is the adjoint method, which enables efficient computation of gradients in PDE-constrained systems by solving an auxiliary adjoint equation \cite{jameson1988aerodynamic,brenner2024variational,strofer2021end}. The adjoint method is particularly advantageous when the number of model inputs (e.g., GNN parameters) far exceeds the number of outputs (e.g., bulk permeability), a common scenario in physics-informed machine learning \cite{martins2021engineering}. In our framework, it allows for efficient training of the GNN by backpropagating sensitivities through the pore network model solver, forming an end-to-end differentiable system that ensures the predicted conductance fields lead to physically consistent and accurate permeability estimates~\cite{strofer2021end}.

\begin{table}[t]
\caption{Analogy between Reynolds-averaged Navier–Stokes (RANS) and Pore Network Model (PNM) frameworks in terms of direct and indirect data.}
\label{tab:rans-pnm-analogy}
\centering
\begin{tabular}{p{0.25\textwidth} p{0.35\textwidth} p{0.35\textwidth}}
\hline
\textbf{Concept} & \textbf{RANS model} & \textbf{Pore Network Model} \\
\hline
Direct data & Labeled Reynolds stress fields & Hydraulic conductance \\
Indirect data & Velocities, drag, lift & Bulk permeability \\
Closure needed for & Reynolds stresses & Hydraulic conductance \\
Physical solver & PDE-based RANS solver & Pore network flow solver \\
\hline
\end{tabular}
\end{table}

The remainder of the paper is organized as follows. In the Materials and Methods section, we begin by introducing the general concept of the proposed framework. We then present the PNM and GNN model, followed by a detailed description of the end-to-end differentiable GNN-embedded pore network model. This is followed by an overview of the dataset used for model training and evaluation. In the Results section, we compare the performance of the pure data-driven GNN model, the traditional PNM, and the proposed hybrid GNN-embedded PNM across different datasets. We also investigate model interpretability through a gradient-based analysis of permeability with respect to node and edge features. Finally, we conclude the paper with a summary of key findings in the Conclusion section.

\section{Materials and Methods}

\begin{figure}[t]
\centering
\includegraphics[width=1.0\textwidth]{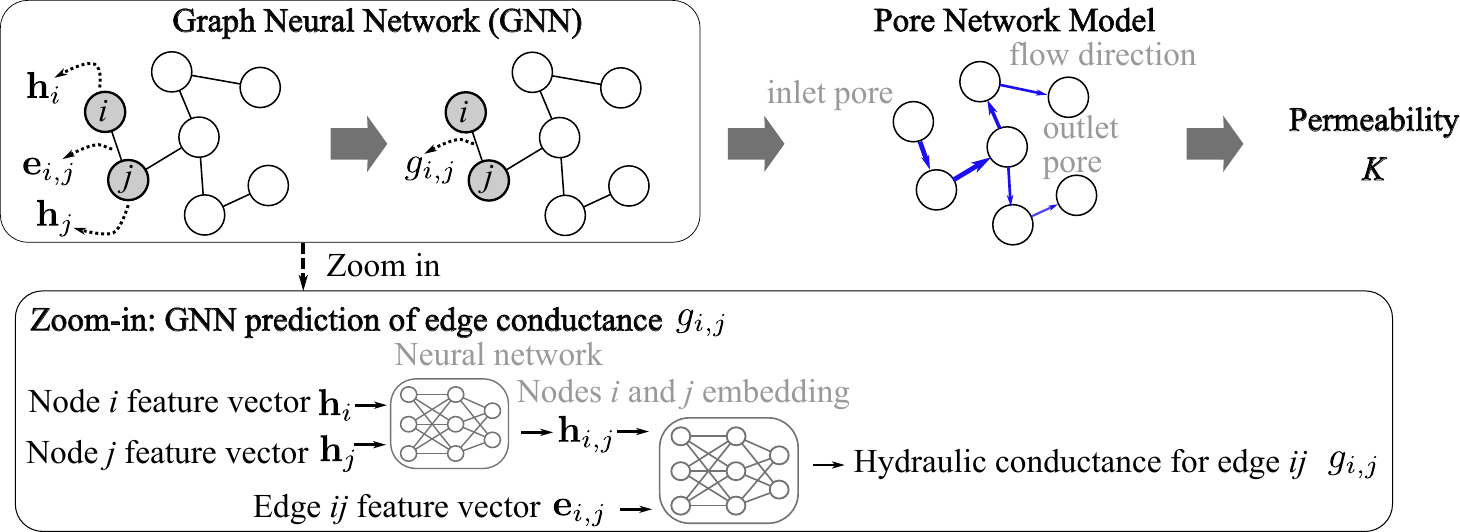}
\caption{
Overview of the proposed graph neural network (GNN)-embedded pore network model (PNM) framework for permeability prediction. 
The GNN takes as input the feature vectors of nodes \( i \) and \( j \) (\( \mathbf{h}_i, \mathbf{h}_j \)) and the edge connecting them (\( \mathbf{e}_{i,j} \)), and outputs the predicted hydraulic conductance \( g_{i,j} \) for each pore-throat pair. These conductance values are passed into a physics-based PNM solver that applies Darcy’s law and mass conservation to compute the bulk permeability \( K \). The bottom box is a zoom-in of the GNN module (top left), illustrating how node and edge features are used to compute the hydraulic conductance \( g_{i,j} \) for each pore-throat pair. Further details on the GNN-embedded pore network model architecture are illustrated in Figure~\ref{fig:framework}. Descriptions of the input node and edge features are provided in Tables~\ref{tab:pore-properties} and \ref{tab:throat-properties}, respectively.
}
\label{fig:gnn-pnm-overview}
\end{figure}

We develop a hybrid modeling framework that integrates GNNs with a physics-based PNM to predict the permeability of porous media. The core idea is to replace analytically derived pore-scale hydraulic conductance values, which are often based on simplified geometric assumptions, with data-driven predictions from a GNN trained on bulk permeability observations. In this framework, the GNN takes the graph representation of a pore network as input and predicts edge-wise hydraulic conductance values $g_{i,j}$ based on the node feature vectors $\mathbf{h}_i$, $\mathbf{h}_j$ and the edge feature vector $\mathbf{e}_{i,j}$ (see Figure~\ref{fig:gnn-pnm-overview}). These predicted conductances are then used in a traditional PNM solver to compute the bulk permeability $K$ by solving a system of equations based on Darcy’s law and mass balance. Crucially, the GNN is trained indirectly, not by comparing its outputs (conductances) to labeled data, but by optimizing its parameters to minimize the mismatch between the PNM solver’s permeability output $K$ and the reference value. This requires backpropagating the loss through the PNM solver, which is made tractable using the adjoint method. The adjoint method efficiently computes the sensitivities of the loss with respect to the predicted conductance values $g_{i,j}$, allowing gradient-based optimization of the GNN weights. This framework combines the physical interpretability and efficiency of pore network modeling with the flexibility of deep learning, enabling permeability prediction from complex pore structures without the need for pore-scale labels.

\begin{figure}
\noindent\includegraphics[width=\textwidth]{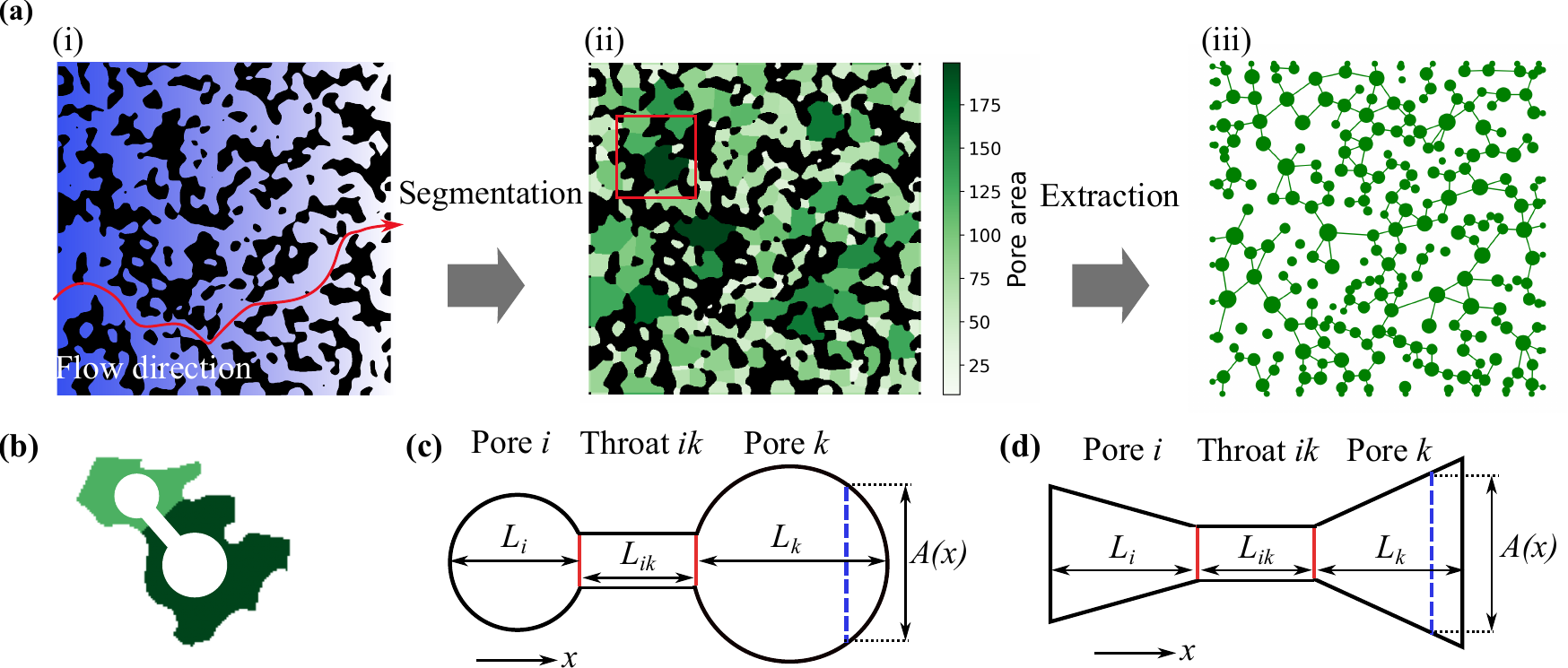}
\caption{
Workflow for pore network extraction and geometric idealization. (a) Workflow for pore network extraction: (i) Preparation of porous media image; (ii) segmentation of the image into distinct pore regions; (iii) extraction of the pore network. Panel (b) shows an example of a pore pair (corresponding to the red rectangle in panel (a.ii)), where the pore shapes are idealized into a standard geometry forms, such as (c) spheres and cylinders, or (d) cones and cylinders, for pores and throats. Here, $L$ denotes length, $A(x)$ is the cross-sectional area, and the subscripts $i$, $k$, and $ik$ refer to pore $i$, pore $k$, and throat $ik$, respectively.}
\label{fig:pnm}
\end{figure}

\subsection{Pore Network Model for Permeability Calculation}
\label{sec:pnm}
The pore network model is a reduced-order representation of porous media that simplifies the pore space into a graph structure, where pores are modeled as nodes and throats as edges. The network is constructed by segmenting a porous media image into distinct regions (Figure~\ref{fig:pnm}a.i-ii), each representing a pore. These segmented regions are then mapped to graph nodes, and their connectivity relationships form the edges of the network (Figure~\ref{fig:pnm}a.ii-iii). Fluid flow through this network is governed by mass conservation at each pore and Darcy’s law along each throat. To simulate flow, each edge in the network must be assigned a hydraulic conductance value, which quantifies how easily fluid can pass through the connecting throat. Given a set of edge-wise conductance values \(\mathbf{g}\), the PNM uses these as coefficients in a system of discrete mass-balance equations. After applying boundary conditions (e.g.\ inlet and outlet pressures), the discrete mass-balance equations can be assembled into a linear system
\begin{linenomath*}
\begin{equation}
\mathbf{A}\left(\mathbf{g}\right)\mathbf{x} = \mathbf{b},
\label{eq:Axb1}
\end{equation}
\end{linenomath*}
where $\mathbf{x}$ is the pore-pressure vector, $\mathbf{b}$ represents the boundary conditions, and $\mathbf{A}$ is a sparse matrix depending on the hydraulic conductance $\mathbf{g}$. Solving Equation~\ref{eq:Axb1} for $\mathbf{x}$ yields the velocity field, from which one computes the net flow rate $Q_{\mathrm{in}}$ at the inlet. Darcy's law then gives the overall permeability $K$ as
\begin{linenomath*}
\begin{equation}
K = c Q_{\mathrm{in}}(\mathbf{x}),
\label{eq:KcQ1}
\end{equation}
\end{linenomath*}
where $c$ is a known factor determined by fluid properties, domain size, and the applied pressure drop. A detailed description of the PNM system matrix assembly and its permeability computation in this work can be found in \ref{app:PNM}. Equations~\ref{eq:Axb1} and~\ref{eq:KcQ1} together define the general pore network model solver for the forward problem.

Therefore, the porous media permeability $K$ is determined by the pore structure and the hydraulic conductivity $g$ for each pore pair. The hydraulic conductivity $g$ is defined by the geometry of the pore-throat-pore conduit which is related to the hydraulic size factor~$F$ by:

\begin{linenomath*}
\begin{equation}
g = \frac{F(L_i, L_{ik}, L_k, A(x), S)}{\mu}
\label{eq:hydraulic_size_factor}
\end{equation}
\end{linenomath*}
where $\mu$ is the fluid dynamic viscosity, $L_i$ and $L_k$ denote the lengths of the two connected pores, $L_{ik}$ is the throat length, $A(x)$ represents the cross-sectional area along the conduit, and $S$ is the shape factor dependent on geometric assumptions. The hydraulic size factor $F$ is computed by integrating geometric properties along the pore–throat–pore conduit. Different geometric assumptions for the conduit segments lead to varying shape factors $S$, which influence the accuracy of the hydraulic conductance $g$. Figure~\ref{fig:pnm}b shows an example of a pore pair region extracted from the segmented domain in Figure~\ref{fig:pnm}a.ii. To extract pore and throat properties from such subdomain images, one must adopt specific assumptions about the shapes of pores and throats. For instance, Figures~\ref{fig:pnm}c and~\ref{fig:pnm}d illustrate two common geometric configurations: in Figure~\ref{fig:pnm}c, pores are modeled as spheres and throats as cylinders, while in Figure~\ref{fig:pnm}d, pores are represented as cones and throats remain cylindrical. The accuracy of hydraulic conductance $g$ predictions strongly depends on the proper estimation of the size factor $F$, which is shape-dependent. Therefore, when calculating pore network properties, we consider a variety of idealized geometries, including pyramids, cuboids, cones, cylinders, trapezoids, rectangles, squares, and cubes. These idealized geometries are then used to approximate pore and throat properties. The geometric properties derived from these assumptions, as summarized in Tables~\ref{tab:pore-properties} and~\ref{tab:throat-properties}, are explicitly used as model inputs due to their strong correlation with hydraulic conductance $g$. This approach provides a rich and diverse set of inputs that enables robust and flexible training of the deep learning model.

\subsection{Graph Neural Network Method}
\label{sec:gnn}

While the PNM provides a physically interpretable and computationally efficient framework for permeability prediction, its accuracy is fundamentally limited by the assumptions used to estimate hydraulic conductance \( g \). As described in Section~\ref{sec:pnm}, the computation of hydraulic conductance \( g \) depends on idealized geometric assumptions used to approximate the pore--throat--pore conduit, which can introduce significant errors when applied to complex and irregular pore structures~\cite{liu2022pore, misaghian2022prediction}. To overcome this limitation, we propose using GNNs to learn a data-driven mapping from structural features of the pore network to permeability-related quantities. Recent work~\cite{zhao2025computationally} demonstrates that a global, graph-level GNN approach can accurately predict permeability. This method represents the entire pore network extracted from a porous media image as a graph, and then applies a GNN to directly estimate the bulk permeability. It is a purely data-driven approach that does not rely on any physical solver. While this method offers significant computational efficiency over traditional numerical approaches, its accuracy may degrade when applied to datasets at different scales, and its lack of embedded physical principles limits interpretability and generalization performance. Details of this pure data-driven GNN model are provided in~\ref{app:gnn}. 

\begin{figure}
\noindent\includegraphics[width=\textwidth]{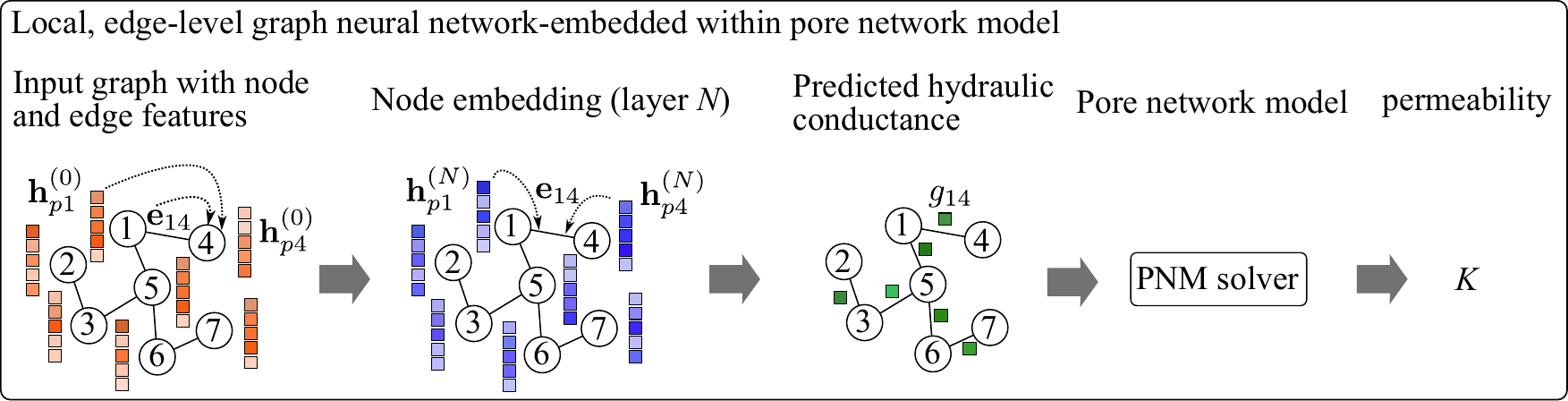}
\caption{
Illustration of the workflow for the graph neural network-embedded pore network model. The model takes as input the initial node features $\mathbf{h}^{(0)}$ and edge features $\mathbf{e}$. After processing through the graph neural network, The updated node embeddings $\mathbf{h}^{(N)}$ along with original edge features $\mathbf{e}$ are used to predict hydraulic conductance $g$ for each edge. For example, $g_{14}$ is predicted using node embeddings $\mathbf{h}_{p1}^{(N)}$, $\mathbf{h}_{p4}^{(N)}$ and edge features $\mathbf{e}_{14}$. The resulting graph with predicted conductances is then passed to a pore network model solver to compute the effective permeability $K$.
}
\label{fig:gnn2}
\end{figure}

To address these shortcomings,  in this work we introduce an approach based on an edge-level GNN, illustrated in Figure~\ref{fig:gnn2}. In this hybrid approach, the GNN is embedded within the pore network solver to predict edge-level hydraulic conductance \( g_{i,j} \) using features of the connected pores and throats. These predicted conductance values are then passed into the PNM solver to compute the bulk permeability. This strategy preserves physical consistency while enabling data-driven learning from indirect supervision using only permeability observations. The GNN-embedded pore network model begins with an input graph containing node features \(\mathbf{h}_p^{(0)}\) and edge features~\(\mathbf{e}_{pp'}\). We iteratively update each node \( p \) and edge~\( e \) across multiple graph neural network layers. Let \(\mathbf{h}_p^{(\ell)}\) denote the embedding of node~\( p \) at layer \( \ell \), and \(\mathbf{e}_{pp'}\) represent the features of the edge connecting node \( p \) to its neighbor~\( p' \). Using a message-passing framework, the intermediate message~\(\mathbf{H}_p^{(\ell+1)}\) is computed by aggregating information from neighboring nodes and edges:
\begin{linenomath*}
\begin{equation}
\mathbf{H}_p^{(\ell+1)} = f_{\mathrm{msg}, \ell+1}\bigl(\mathbf{h}_{p'}^{(\ell)},\, \mathbf{e}_{pp'}\bigr),
\label{eq:node_message}
\end{equation}
\end{linenomath*}
where \(f_{\mathrm{msg}, \ell+1}\) is a learnable function incorporating the neighbor embedding \(\mathbf{h}_{p'}^{(\ell)}\) and the corresponding edge features \(\mathbf{e}_{pp'}\). The updated embedding of node \( p \) at the next layer is then computed by applying a learnable update function \( f_{\mathrm{apply}, \ell+1} \), which combines the node’s current embedding \( \mathbf{h}_p^{(\ell)} \) with the aggregated message \( \mathbf{H}_p^{(\ell+1)} \):
\begin{linenomath*}
\begin{equation}
\mathbf{h}_p^{(\ell+1)} = f_{\mathrm{apply}, \ell+1}\bigl(\mathbf{h}_p^{(\ell)},\, \mathbf{H}_p^{(\ell+1)}\bigr).
\label{eq:node_update}
\end{equation}
\end{linenomath*}
After a series of message-passing layers \(f_{\mathrm{msg}}\) and \( f_{\mathrm{apply}} \), node embeddings \(\mathbf{h}_{p}\) are iteratively updated to yield the final representations \(\mathbf{h}_p^{(N)}\). Following the last message-passing layer, the edge-wise hydraulic conductance~\(g_{pp'}\) is predicted using a multilayer perceptron applied to the concatenated features of the connected node embeddings and the original edge attributes:
\begin{linenomath*}
\begin{equation}
\mathbf{m}_{pp'} = f_{\mathrm{hid}}\left([\mathbf{h}_p^{(N)};\, \mathbf{h}_{p'}^{(N)};\, \mathbf{e}_{pp'}]\right),
\label{eq:edge_hidden}
\end{equation}
\end{linenomath*}
\begin{linenomath*}
\begin{equation}
g_{pp'} = f_{\mathrm{out}}(\mathbf{m}_{pp'}),
\label{eq:edge_output}
\end{equation}
\end{linenomath*}
where \(f_{\mathrm{hid}}\) and \(f_{\mathrm{out}}\) are multilayer perceptron layers with ReLU and Softplus activations, respectively; $\mathbf{m}_{pp'}$ is the intermediate edge representation. The detailed architecture of the edge-level GNN is provided in Table~\ref{tab:gnn-architecture2}. The predicted hydraulic conductance~\(g\) from the edge-level GNN is then assembled into the matrix~$\mathbf{A}$ in Equation~\ref{eq:Axb1}, which is used to solve the pore network model for permeability computation.

\subsection{GNN-Embedded Pore Network Model and Gradient Computation}
\label{sec:gnn-pnm}

\begin{figure}
\noindent\includegraphics[width=1.0\textwidth]{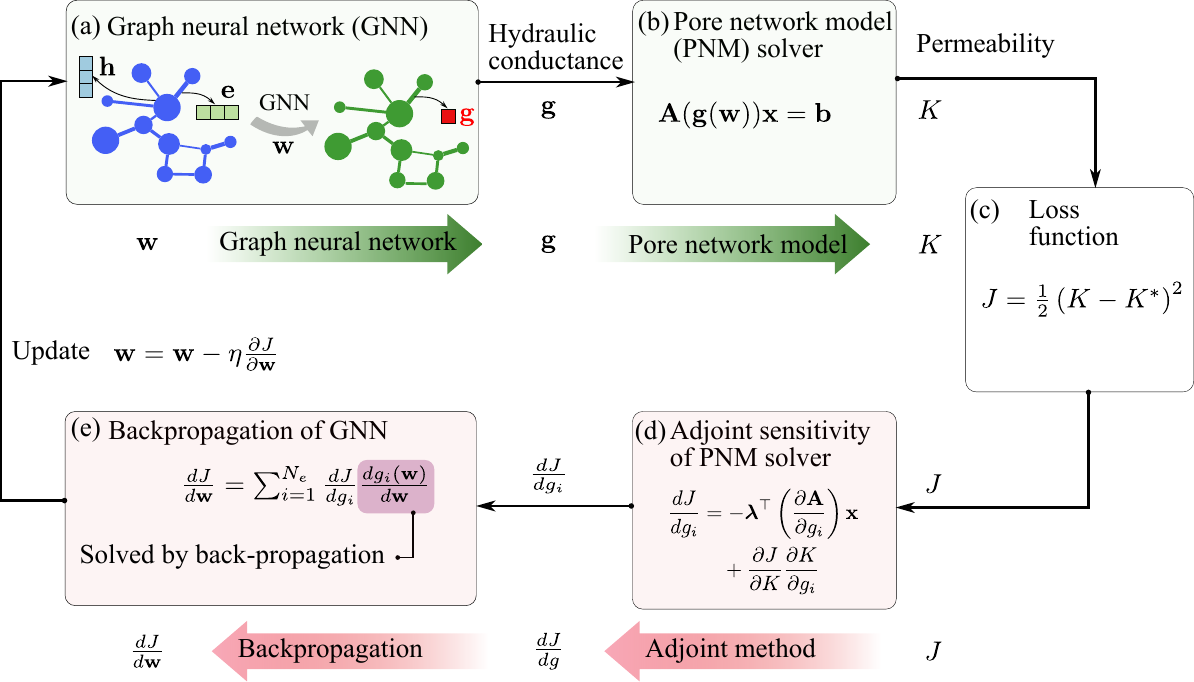}
\caption{
End-to-end differentiable framework of the graph neural network (GNN)-embedded pore network model (PNM). 
Forward process: (a) A GNN with weights \( \mathbf{w} \) predicts edge-wise hydraulic conductance values \( \mathbf{g}(\mathbf{w}) \) from node features \( \mathbf{h} \) and edge features \( \mathbf{e} \) of the pore network graph. 
(b) The predicted hydraulic conductances \( \mathbf{g}(\mathbf{w}) \) are passed to a PNM solver to compute permeability \( K \). 
After the forward process: (c) The loss function compares the predicted permeability \( K \) to a reference value \( K^* \), defining the loss \( J \). 
Backward process: (d) The discrete adjoint method applied to the PNM computes the gradient of the loss with respect to hydraulic conductance \( d J / d g_i \). 
(e) Backpropagation applies the chain rule to compute the total gradient of the loss with respect to GNN weights \( d J / d \mathbf{w} \) by combining adjoint-based gradients of the loss with respect to hydraulic conductance \( d J / d g_i \) and GNN-derived gradients of hydraulic conductance with respect to GNN weights \( d g_i(\mathbf{w}) / d \mathbf{w} \).
After the backward process: The GNN weights \( \mathbf{w} \) are updated according to the gradient descent rule with a learning rate \( \eta \).
}

\label{fig:framework}
\end{figure}

The edge-level GNN predicts pore-scale hydraulic conductance values \( g_{i,j} \) based on local pore and throat features. These predictions are then used by PNM to compute the bulk permeability. However, a key challenge arises: direct training data for hydraulic conductance at the pore scale are generally unavailable, as such quantities cannot be measured experimentally and are costly to obtain from high-fidelity simulations. This makes direct supervision of the GNN infeasible. To address this challenge, we propose a hybrid, physics-informed training framework that integrates the GNN with the PNM solver in an end-to-end differentiable manner. The workflow of the end-to-end differentiable framework for the GNN-embedded pore network model is illustrated in Figure~\ref{fig:framework}.

\subsubsection{Forward Model and Loss Function}

The end-to-end differentiable framework begins with the forward process, which predicts the bulk permeability of a porous medium based on its graph representation. This process combines a graph neural network that estimates edge-wise hydraulic conductances from node and edge features, with a pore network model solver that computes the resulting permeability. As illustrated in Figure~\ref{fig:framework}a--c, the forward model is followed by a loss function that compares the predicted permeability to a reference value for training purposes. These steps involve the following key computational components:

\begin{enumerate}
  \item \textbf{Hydraulic Conductance Prediction via GNN.}  
  Given node features $\mathbf{h}$ and edge features $\mathbf{e}$, the GNN with learnable parameters $\mathbf{w}$ predicts the hydraulic conductance for each edge in the pore network graph (Figure~\ref{fig:framework}a):
  \begin{linenomath*}
  \begin{equation}
  \mathbf{g} = \mathrm{GNN}(\mathbf{h}, \mathbf{e}; \mathbf{w}).
  \end{equation}
  \end{linenomath*}
  This prediction is computed using the message passing and edge output rules previously defined in Equations~\ref{eq:node_message}--\ref{eq:node_update} and Equation~\ref{eq:edge_output}, respectively.

  \item \textbf{Permeability Calculation via PNM Solver.}  
  The predicted conductance values $\mathbf{g}(\mathbf{w})$ are passed to the PNM solver to compute the bulk permeability $K$:
  \begin{linenomath*}
  \begin{equation}
  K = \mathrm{PNM}(\mathbf{g}).
  \end{equation}
  \end{linenomath*}
  These steps correspond to Figure~\ref{fig:framework}b and are defined in detail in Equations~\ref{eq:Axb1} and~\ref{eq:KcQ1}.
\end{enumerate}

\noindent
 
Once the permeability $K$ is computed, it is compared against a ground-truth permeability $K^*$ (e.g., from lattice Boltzmann simulations). The resulting loss function is defined as:
\begin{linenomath*}
\begin{equation}
J = \frac{1}{2}\left(K - K^*\right)^2,
\label{eq:objfunc}
\end{equation}
\end{linenomath*}
which penalizes the discrepancy between the predicted and reference permeability values (Figure~\ref{fig:framework}c). This process establishes the differentiable path from GNN parameters~$\mathbf{w}$ to the loss $J$, enabling gradient-based training in the subsequent backward step.

\subsubsection{Gradient Computation with Backpropagation and Adjoint Method}

Following the loss function, the end-to-end differentiable framework is completed by the backward process, which enables learning of the GNN parameters \(\mathbf{w}\) based on the loss computed from the predicted permeability (Figure~\ref{fig:framework}d--e). This is achieved through gradient-based optimization, where the weights are updated using the computed gradients of the loss function. The update rule follows standard gradient descent:
\begin{linenomath*}
\begin{equation}
\mathbf{w} = \mathbf{w} - \eta \, \frac{d J}{d \mathbf{w}},
\label{eq:weight-update}
\end{equation}
\end{linenomath*}
where $\eta$ is the learning rate. The total gradient of the loss with respect to GNN weights~\(\frac{dJ}{d\mathbf{w}}\) is obtained using the chain rule:
\begin{linenomath*}
\begin{equation}
\frac{d J}{d \mathbf{w}} = \sum_{i=1}^{N_e} \frac{d J}{d g_i} \cdot \frac{d g_i(\mathbf{w})}{d \mathbf{w}},
\label{eq:chain-rule-general}
\end{equation}
\end{linenomath*}
where \(g_i\) is the predicted hydraulic conductance for edge \(i\). This gradient consists of two key components that correspond to the GNN and PNM stages of the forward model: the sensitivity of the predicted conductances with respect to the GNN weights, \(\tfrac{d g_i}{d \mathbf{w}}\), computed via automatic differentiation, and the sensitivity of the loss with respect to each conductance, \(\tfrac{d J}{d g_i}\), obtained using the discrete adjoint method. These components are detailed below:

\begin{enumerate}
  \item \textbf{Gradient of GNN Output with Respect to Weights (\(\mathbf{d g_i / d \mathbf{w}}\)).}  
  This term corresponds to the first step of the forward model, where the GNN predicts edge-wise conductances \(\mathbf{g} = \mathrm{GNN}(\mathbf{h}, \mathbf{e}; \mathbf{w})\). During backpropagation, we compute how each predicted conductance \(g_i\) changes with respect to the GNN weights \(\mathbf{w}\), yielding the derivative of hydraulic conductance with respect to GNN weights \(\frac{d g_i(\mathbf{w})}{d \mathbf{w}}\).

  This quantity is obtained using automatic differentiation (AD) through the GNN’s architecture. During the forward pass, each operation, such as message passing, feature updates, and MLP evaluation, is tracked by the automatic differentiation engine. When gradients are needed, automatic differentiation applies the chain rule in reverse through all layers and operations to compute the gradient of hydraulic conductance with respect to GNN weights~\(\frac{d g_i}{d \mathbf{w}}\) efficiently. This approach eliminates the need for manual gradient derivation and supports complex GNN architectures (Figure~\ref{fig:framework}e).

  \item \textbf{Gradient of Loss with Respect to Conductance (\(\mathbf{d J / d g_i}\)).}  
  This term corresponds to the second step of the forward model, where the PNM solver takes predicted conductances~\(\mathbf{g}\) and computes the permeability \(K = \mathrm{PNM}(\mathbf{g})\). To train the model, we must compute how the loss \(J\) changes with respect to each predicted conductance \(g_i\), yielding the derivative \(\frac{d J}{d g_i}\).

  This influence is not straightforward because each \(g_i\) affects the pressure solution~\(\mathbf{x}\) through the pore network linear system \(\mathbf{A}(\mathbf{g})\, \mathbf{x} = \mathbf{b}\), and the solution \(\mathbf{x}\) subsequently determines the flow rate and permeability \(K\), which enters the loss function. Directly computing the gradient of the loss with respect to hydraulic conductance~\(\frac{d J}{d g_i}\) via finite differences or full backpropagation through the linear solver is inefficient and unstable. Instead, we use the discrete adjoint method (Figure~\ref{fig:framework}d), which introduces an auxiliary linear system:
  \begin{linenomath*}
  \begin{equation}
  \mathbf{A}^\top(\mathbf{g})\, \boldsymbol{\lambda} = \frac{\partial J}{\partial \mathbf{x}},
  \label{eq:adjoint-sys}
  \end{equation}
  \end{linenomath*}
  where \(\boldsymbol{\lambda}\) is the adjoint vector. This system mirrors the structure of the original forward problem but propagates the physical sensitivity of the loss back to each state variable. Once both systems (\(\mathbf{A} \mathbf{x} = \mathbf{b}\) for the forward pass and \(\mathbf{A}^\top \boldsymbol{\lambda} = \partial J/\partial \mathbf{x}\) for the adjoint) are solved, we can compute the total sensitivity of the loss with respect to each conductance \(g_i\) via:
  \begin{linenomath*}
  \begin{equation}
  \frac{d J}{d g_i} = -\boldsymbol{\lambda}^{\mathrm{T}} \left(\frac{\partial \mathbf{A}}{\partial g_i}\right) \mathbf{x} + \frac{\partial J}{\partial K} \cdot \frac{\partial K}{\partial g_i},
  \label{eq:dJdgi-general}
  \end{equation}
  \end{linenomath*}

  Since the loss function is defined in Equation~\ref{eq:objfunc}, its derivative with respect to permeability is:
  \begin{linenomath*}
  \begin{equation}
  \frac{\partial J}{\partial K} = K - K^*.
  \label{eq:dJdK}
  \end{equation}
  \end{linenomath*}

  This adjoint-based formulation is both memory- and computation-efficient, as it avoids repeated linear solves for each \(g_i\) and naturally incorporates the sparse structure of the system matrix \(\mathbf{A}\). A detailed derivation of the discrete adjoint system and the sensitivity of the loss with respect to conductance, \(\tfrac{d J}{d g_i}\), is provided in~\ref{app:adjoint}.
\end{enumerate}

Together, these two components yield the full gradient of the loss with respect to GNN weights~\(\frac{dJ}{d\mathbf{w}}\), enabling accurate and physically consistent updates to the GNN parameters via Equation~\ref{eq:weight-update}. The detailed steps for the end-to-end training of the GNN-embedded pore network model are summarized in Table~\ref{tab:algorithm1}.

\begin{table}[t]
\caption{Algorithm 1: End-to-end training of the GNN-embedded pore network model}
\label{tab:algorithm1}
\centering
\begin{tabular}{p{0.97\textwidth}}
\hline
\textbf{Input:} Node features $\mathbf{X}$, edge features $\mathbf{E}$, boundary conditions, ground-truth permeability $K^*$, learning rate $\eta$. \\
\textbf{Output:} Trained GNN parameters $\mathbf{w}$. \\
\textbf{Initialize:} Randomly initialize GNN parameters $\mathbf{w}$. \\
\textbf{for} epoch = $1$ to num\_epochs \textbf{do}: \\
1. \textbf{GNN Forward Pass}:\\
~~~Compute throat conductances $\{g_i(\mathbf{w})\}$ via message-passing.\\
2. \textbf{PNM Solve}:\\
~~~Assemble system matrix $\mathbf{A}(\mathbf{g}(\mathbf{w}))$.\\
~~~Solve linear system $\mathbf{A}\,\mathbf{x}=\mathbf{b}$ for pressures $\mathbf{x}$.\\
3. \textbf{Compute Loss}:\\
~~~Compute permeability $K = c Q_{\mathrm{in}}(\mathbf{x})$.\\
~~~Evaluate loss $J=\frac{1}{2}(K - K^*)^2$.\\
4. \textbf{Adjoint Solve}:\\
~~~Compute gradient w.r.t.~$\mathbf{x}$: $\frac{\partial J}{\partial \mathbf{x}}=(K-K^*) c \frac{\partial Q_{\mathrm{in}}}{\partial \mathbf{x}}$.\\
~~~Solve adjoint system: $\mathbf{A}^{\mathrm{T}} \mathbf{\lambda}=(\frac{\partial J}{\partial \mathbf{x}})^T$.\\
5. \textbf{Compute Physical Gradient}:\\
~~~For each throat $i$, compute gradient of the loss with respect to hydraulic conductance $\frac{d J}{d g_i}$ using Equations~\ref{eq:dJdgi-general} or \ref{eq:dJdgi-final-app}.\\
6. \textbf{Backpropagation}:\\
~~~Compute gradients of the loss with respect to GNN weights: $\frac{d J}{d \mathbf{w}} = \frac{d J}{d g_i}\frac{d g_i}{d \mathbf{w}}$ (via automatic differentiation).\\
7. \textbf{Update}:\\
~~~Update parameters: $\mathbf{w} \leftarrow \mathbf{w} - \eta \frac{d J}{d \mathbf{w}}$.\\
\textbf{end for}\\
\hline
\end{tabular}
\end{table}

\subsection{Dataset}

\begin{figure}
\noindent\includegraphics[width=1.0\textwidth]{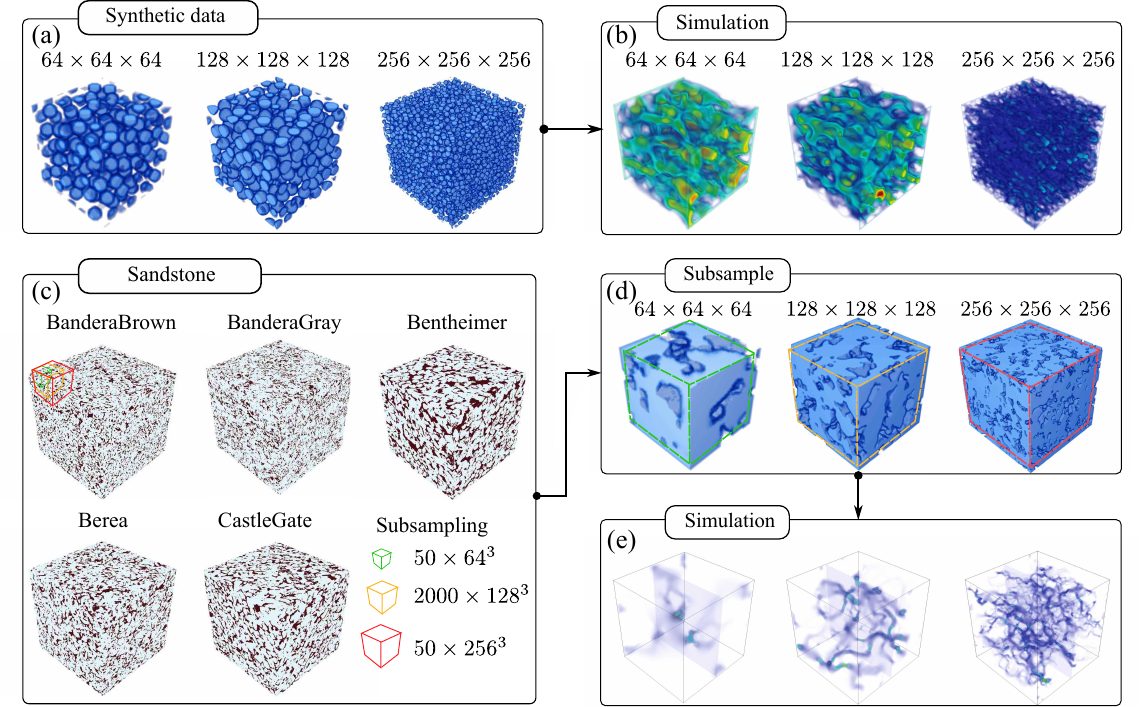}
\caption{Overview of the dataset used for training and testing the deep learning model for permeability prediction.
(a) Synthetic porous media generated using PoreSpy with random sphere packing at three resolutions: $64^3$, $128^3$, and $256^3$ voxels.
(b) Flow simulation results on the corresponding synthetic geometries.
(c) High-resolution sandstone images from the Digital Rocks Portal, including Bandera Brown, Bandera Gray, Bentheimer, Berea, and Castle Gate formations.
(d) Subsamples extracted from the sandstone images at three resolutions: 50 samples of $64^3$ voxels (green boxes), 2000 samples of $128^3$ voxels (orange boxes), and 50 samples of $256^3$ voxels (red boxes).
(e) Flow simulation results for the subsampled sandstone images.}
\label{fig:dataset}
\end{figure}

To evaluate the performance of our proposed framework in comparison with traditional PNM and standalone GNN models, we constructed a comprehensive dataset consisting of both synthetic and real sandstone samples. The synthetic samples were generated procedurally, while the real sandstone samples were obtained from high-resolution X-ray CT scans. An overview of the dataset is provided in Figure~\ref{fig:dataset}.

The synthetic porous media were generated using the open-source software PoreSpy \cite{gostick2019porespy}, creating random sphere packings to simulate porous structures. These synthetic datasets were generated at three different resolutions: 2000 samples with dimensions of $128^3$ voxels, 50 samples with dimensions of $64^3$ voxels, and another 50 samples with dimensions of $256^3$ voxels. The grain size distributions were consistent across all three resolutions; thus, as the image dimension increased from 64 to 256 voxels, the number of grains increased exponentially (Figure~\ref{fig:dataset}a). From the synthetic dataset, 90\% of the $128^3$ voxel images were allocated to training the model, while the remaining 10\% of the $128^3$ voxel images, along with all $64^3$ and $256^3$ voxel images, were reserved for testing the trained model. 

In addition to synthetic samples, we included real sandstone images sourced from the Digital Rocks Portal \cite{prodanovic2023digital}. The sandstone types incorporated in our study include Berea, Bandera Brown, Bandera Gray, Bentheimer, and Castle Gate. Initially, these sandstone samples had dimensions of $1000 \times 1000 \times 1000$ voxels (Figure~\ref{fig:dataset}c). These large images were then subdivided into 2000 smaller sub-samples of $128^3$ voxels. Additionally, 50 subsamples each of $64^3$ and $256^3$ voxel dimensions were also extracted for testing purposes (Figure~\ref{fig:dataset}d).

All synthetic and real sandstone images underwent permeability calculation using the lattice Boltzmann method (LBM) implemented in MPLBM-UT (Figures~\ref{fig:dataset}b and e). LBM is a computational fluid dynamics (CFD) technique based on mesoscopic fluid models, effectively capturing intricate geometrical features and fluid behaviors within porous structures. In LBM, the fluid domain is discretized into lattice grids, where fluid dynamics are modeled by the evolution of a particle distribution function, incorporating collision and streaming processes. More detailed information about the LBM computation can be found in the MPLBM-UT documentation \cite{santos2022mplbm}.

Following the image preparation and LBM permeability simulation, each porous structure image was processed using PoreSpy to extract its pore network. To achieve this, a subnetwork of the oversegmented watershed (SNOW) algorithm is applied \cite{gostick2019porespy}. The SNOW algorithm uses a marker-based watershed segmentation approach to partition the pore space of the images (Figure~\ref{fig:pnm}a) into distinct pore regions (Figure~\ref{fig:pnm}b). Specifically, the algorithm first computes a distance transform of the binary image, finding the center of each pore. Then, a Gaussian filter is applied to reduce spurious peaks, ensuring accurate pore center identification. Finally, the marker-based watershed segmentation partitions the image into individual pore regions, assigning each voxel to a distinct pore. The partitioned image is then processed using OpenPNM \cite{gostick2016openpnm}, generating a detailed numerical description of the pore network, including properties for each pore and connecting throat. The pore networks were subsequently analyzed to calculate permeability (pore network model-derived permeability). Additionally, these pore networks were represented as graph structures, with nodes representing pores and edges representing throats (Figure~\ref{fig:dataset}c). Detailed geometric and topological attributes were assigned to nodes and edges, comprising 8 pore properties such as pore diameter, coordination number, surface area and volume and 15 throat properties such as throat diameter, length, cross-sectional area, and hydraulic size factors. A comprehensive list and detailed explanations of these pore and throat properties are provided in \ref{app:features}.

To ensure consistent numerical scaling across features with varying physical units and magnitudes, all node and edge features were standardized prior to training. Given a feature \( f \), the normalized value \( z \) is computed using z-score normalization:
\begin{linenomath*}
\begin{equation}
z = \frac{f - \bar{f}}{\sigma}
\label{eq:zscore_normalization}
\end{equation}
\end{linenomath*}
where \( \bar{f} \) and \( \sigma \) denote the global mean and standard deviation of that feature, computed across all samples in the training dataset. This transformation yields dimensionless features \( z \) with zero mean and unit variance. These standardized node and edge features serve as inputs to the graph neural network, ensuring balanced feature contributions and improving training stability.

\section{Results} 

\subsection{Model Performance Comparison}

We compare the predictive performance of three models: a traditional pore network model, a pure data-driven graph neural network model, and the proposed GNN-embedded pore network model in this study. The pore network model directly predicts permeability based on geometric factors derived from pore network attributes. The pure data-driven graph neural network model aggregates pore and throat features via graph convolutions and predicts permeability as a single scalar. The GNN-embedded pore network model integrates these two approaches, using a graph neural network to predict throat-level hydraulic conductance, which is subsequently used in a physics-based pore network solver for permeability computations.

We evaluated permeability predictions on synthetic sphere‐packing images of sizes $64^3$, $128^3$, and $256^3$ voxels. As described in the Methods, only the $128^3$ images were used to train the graph neural network and the GNN-embedded pore network model, while the $64^3$ and $256^3$ images were reserved solely for testing extrapolation capability to unseen scales. In contrast, the pore network model is an analytical pore network computation that does not require training.

The model performances were evaluated using four metrics: mean absolute error (MAE), root mean squared error (RMSE), mean absolute percentage error (MAPE), and coefficient of determination ($R^2$), defined as:

\begin{linenomath*}
\begin{equation}
\text{MAE} = \frac{1}{n}\sum_{i=1}^{n}|y_i - \hat{y}_i|,
\end{equation}
\end{linenomath*}

\begin{linenomath*}
\begin{equation}
\text{RMSE} = \sqrt{\frac{1}{n}\sum_{i=1}^{n}(y_i - \hat{y}_i)^2},
\end{equation}
\end{linenomath*}

\begin{linenomath*}
\begin{equation}
\text{MAPE} = \frac{100\%}{n}\sum_{i=1}^{n}\left|\frac{y_i - \hat{y}_i}{y_i}\right|,
\end{equation}
\end{linenomath*}

\begin{linenomath*}
\begin{equation}
R^2 = 1 - \frac{\sum_{i=1}^{n}(y_i - \hat{y}_i)^2}{\sum_{i=1}^{n}(y_i - \bar{y})^2},
\end{equation}
\end{linenomath*}
where $y_i$ represents observed permeability, $\hat{y}_i$ the predicted permeability, and $\bar{y}$ the mean of observed permeability values.

MAE and RMSE both measure the average magnitude of prediction errors. MAE is the mean of absolute differences, making it easy to interpret in the same units as permeability; lower values indicate more accurate predictions. RMSE penalizes larger errors more strongly due to the squaring operation, which makes it more sensitive to outliers; here as well, lower values correspond to better performance. MAPE expresses the average error as a percentage of the true values, providing a scale-independent measure that facilitates comparison across datasets; smaller percentages reflect higher predictive accuracy. Finally, $R^2$ quantifies the proportion of variance in the observed data that is explained by the model. Values closer to 1 indicate stronger explanatory power, while values near 0 suggest poor predictive ability. In rare cases, very high $R^2$ values may indicate overfitting rather than true generalization.

\subsubsection{Results on Synthetic Data}
\begin{figure}
\noindent\includegraphics[width=\textwidth]{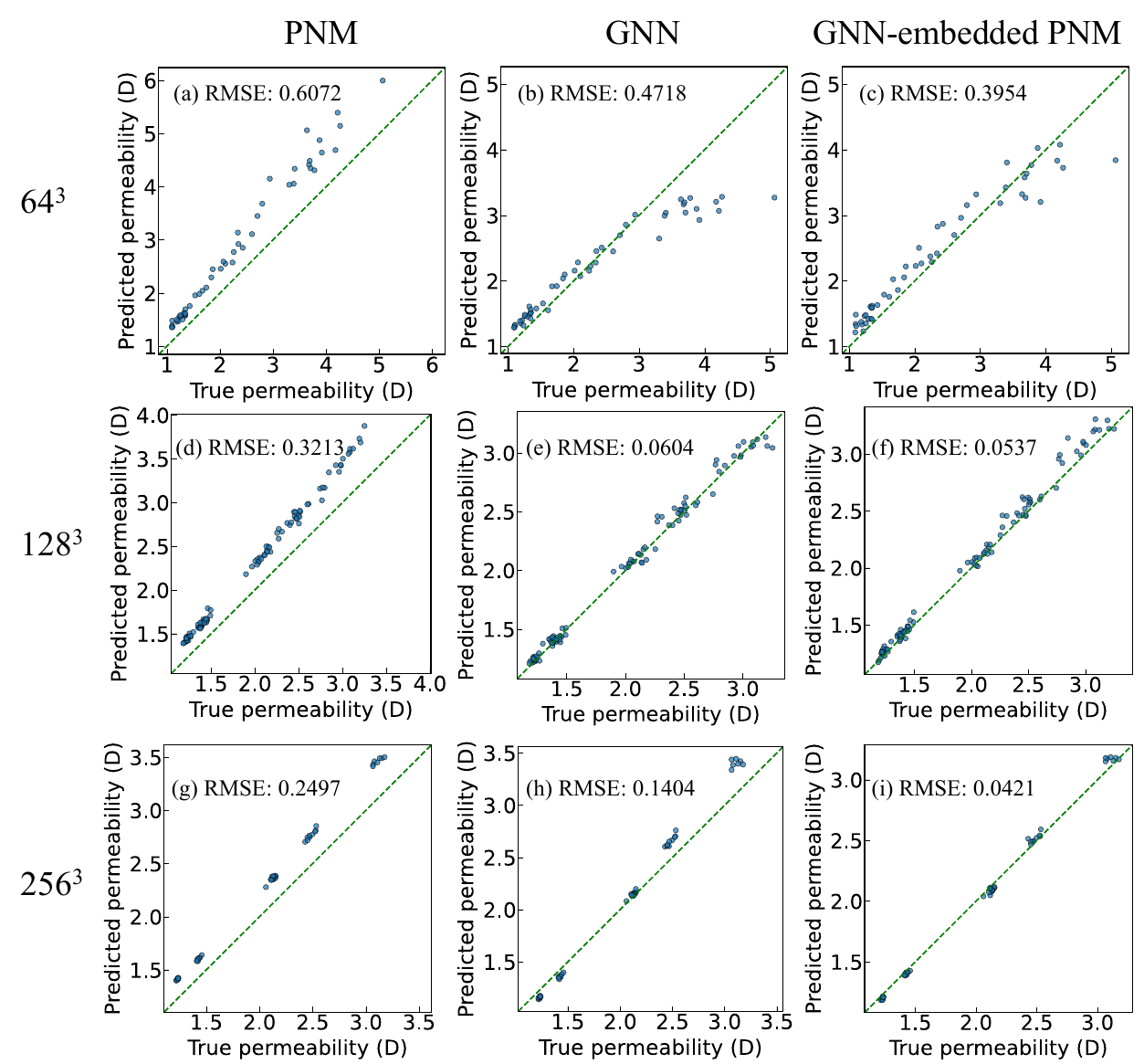}
\caption{Comparison of predicted and true permeability for synthetic datasets at different resolutions using the pore network model (PNM), the pure data-driven graph neural network (GNN) model~\cite{zhao2025computationally}, and the GNN-embedded PNM: (a) PNM at $64^3$, (b) GNN at $64^3$, (c) GNN-embedded PNM at $64^3$, (d) PNM at $128^3$, (e) GNN at $128^3$, (f) GNN-embedded PNM at $128^3$, (g) PNM at $256^3$, (h) GNN at $256^3$, and (i) GNN-embedded PNM at $256^3$.}
\label{fig:synthetic_results}
\end{figure}

In the synthetic datasets, the GNN-embedded PNM consistently outperforms both the traditional pore network model and the pure data-driven GNN, achieving superior accuracy and generalization across varying resolution. Representative scatter plots of predicted versus true permeability for the $64^3$, $128^3$, and $256^3$ synthetic datasets are shown in Figure~\ref{fig:synthetic_results}. Each row corresponds to a specific resolution, while each column compares a different model: PNM (left), GNN (middle), and GNN-embedded PNM (right). In all subplots, the dashed green line represents the ideal 1:1 prediction.

The pore network model exhibits a consistent overestimation bias, particularly at lower resolutions, but maintains stable performance across scales. This bias likely arises from the simplified geometric assumptions used in the analytical hydraulic conductance calculations, which fail to fully capture pore curvature and connectivity. Despite this, the pore network model’s performance remains relatively stable across resolutions (Figures~\ref{fig:synthetic_results}a, d, g). As shown in Table~\ref{tab:synthetic_metrics}, the root mean squared error (RMSE) improves from resolution $64^3$ to $256^3$, indicating modest accuracy gains with increased resolution.

The pure data-driven GNN demonstrates strong performance at the training resolution but fails to generalize across scales. Trained on $128^3$ samples, the model performs well on the same resolution (Figure~\ref{fig:synthetic_results}e). However, its performance drops on the unseen $64^3$ and $256^3$ datasets (Figures~\ref{fig:synthetic_results}b and h). This degradation likely results from the pure data-driven graph neural network’s use of global pooling operations, which aggregate node and edge features into a single graph-level representation. While these operations ensure permutation invariance, they can also obscure critical structural differences between graphs of varying size and connectivity. As a result, the pure data-driven graph neural network struggles to generalize when applied to datasets with different scales.

The GNN-embedded PNM achieves both high accuracy and strong generalization across all resolutions. As shown in Figures~\ref{fig:synthetic_results}c, f, i, the GNN-embedded PNM  consistently achieves low RMSE across all resolutions. On the training resolution ($128^3$), its performance is comparable to the pure data-driven GNN (Figures~\ref{fig:synthetic_results}e and f). However, the key advantage of the GNN-embedded PNM lies in its ability to generalize. On both coarser and finer graphs unseen during training, it significantly outperforms the GNN (Figures~\ref{fig:synthetic_results}c and i). This robustness stems from embedding learned hydraulic conductance values within a physically consistent flow solver. While the GNN captures local geometric dependencies, the PNM maintains fidelity to pore network physics. This combination allows the GNN-embedded PNM to extrapolate across scales more effectively, reducing errors introduced by overfitting or resolution mismatch. In addition to RMSE, other evaluation metrics, including MAE, MAPE, and $R^2$, are also reported in Table~\ref{tab:synthetic_metrics}. The trends observed across these metrics are consistent, further confirming the improved accuracy and generalization of the GNN-embedded PNM model compared to both the pure data-driven GNN and the traditional PNM.

\begin{table}[t]
\caption{Model evaluation metrics for synthetic datasets at different resolutions using the pore network model (PNM), the pure data-driven graph neural network (GNN) model, and the GNN-embedded PNM. The metrics include mean absolute error (MAE), root mean squared error (RMSE), mean absolute percentage error (MAPE), and coefficient of determination ($R^2$).}
\label{tab:synthetic_metrics}
\centering
\begin{tabular}{p{0.15\textwidth} p{0.25\textwidth} p{0.1\textwidth} p{0.1\textwidth} p{0.1\textwidth} p{0.1\textwidth}}
\hline
\textbf{Resolution} & \textbf{Model} & \textbf{MAE} & \textbf{RMSE} & \textbf{MAPE (\%)} & \textbf{$R^2$} \\[4pt]
\hline
$64^3$ & PNM & 0.5361 & 0.6072 & 23.45 & 0.6946 \\[4pt]
   & GNN & 0.3213 & 0.4718 & 12.79 & 0.8156 \\[4pt]
   & GNN-embedded PNM & 0.2841 & 0.3954 & 9.28 & 0.9331 \\[6pt]
\hline
$128^3$ & PNM & 0.3088 & 0.3213 & 15.67 & 0.7544 \\[4pt]
    & GNN & 0.0433 & 0.0604 & 2.16 & 0.9913 \\[4pt]
    & GNN-embedded PNM & 0.0394 & 0.0537 & 1.97 & 0.9931 \\[6pt]
\hline
$256^3$ & PNM & 0.2396 & 0.2497 & 12.18 & 0.8464 \\[4pt]
    & GNN & 0.1045 & 0.1404 & 4.85 & 0.9504 \\[4pt]
    & GNN-embedded PNM & 0.0353 & 0.0421 & 1.80 & 0.9955 \\[6pt]
\hline
\end{tabular}
\end{table}

\subsubsection{Results on Real Sandstone Data}

\begin{figure}
\centering
\includegraphics[width=\textwidth]{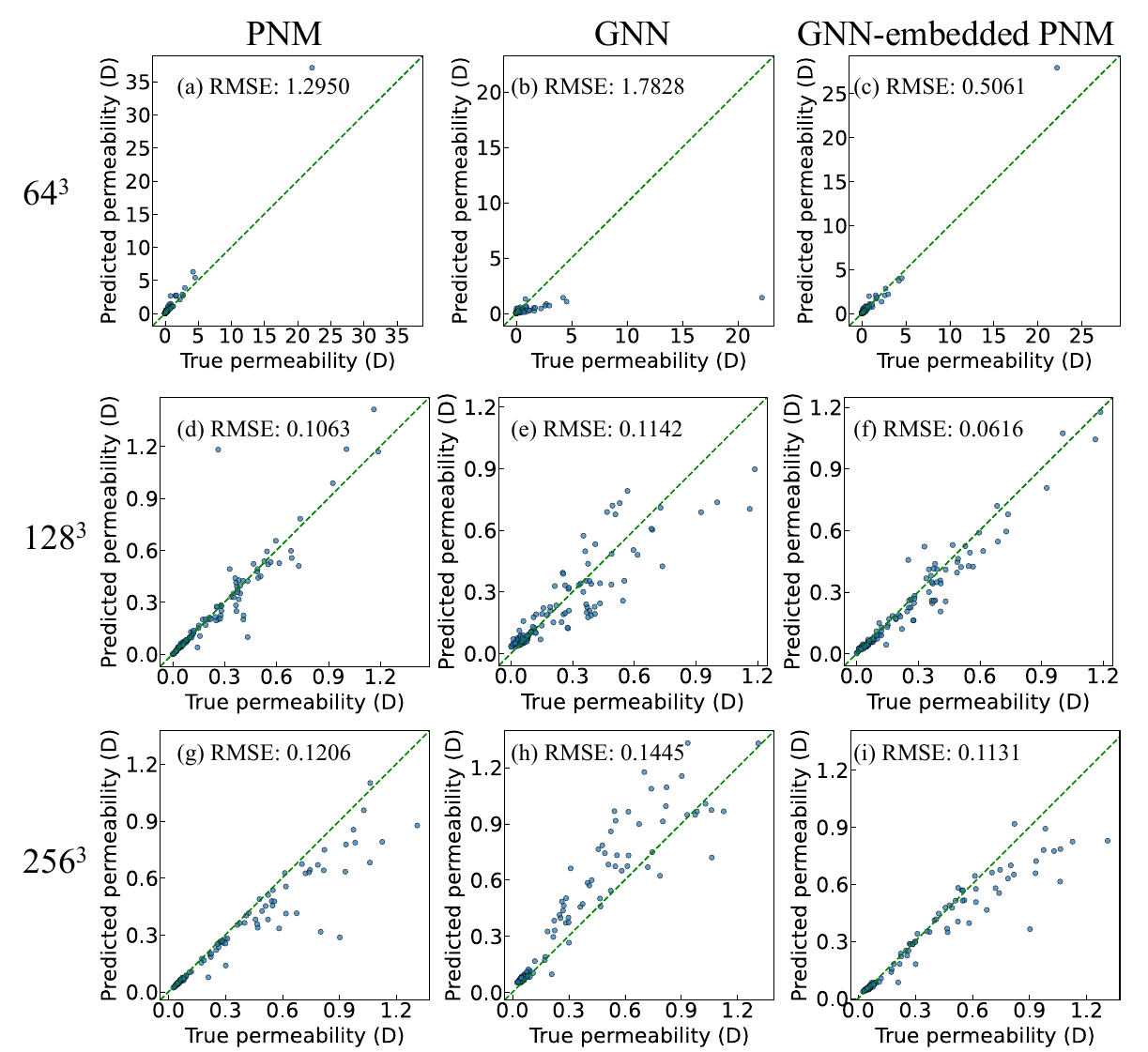}
\caption{Comparison of predicted and true permeability for sandstone datasets at different resolutions using the pore network model (PNM), the pure data-driven graph neural network (GNN) model~\cite{zhao2025computationally}, and the GNN-embedded PNM: (a) PNM at $64^3$, (b) GNN at $64^3$, (c) GNN-embedded PNM at $64^3$, (d) PNM at $128^3$, (e) GNN at $128^3$, (f) GNN-embedded PNM at $128^3$, (g) PNM at $256^3$, (h) GNN at $256^3$, and (i) GNN-embedded PNM at $256^3$.}
\label{fig:sandstone_results}
\end{figure}

On real sandstone datasets, the GNN-embedded PNM again demonstrates superior generalization and accuracy across resolutions, outperforming both the traditional PNM and the pure data-driven GNN. Figure~\ref{fig:sandstone_results} shows the predicted versus true permeability for real sandstone images across three resolutions: $64^3$, $128^3$, and $256^3$. As in the synthetic case, we compare PNM, GNN, and GNN-embedded PNM under the condition that only $128^3$ data were used for training the two GNN-based models.

The PNM predictions appear less biased on the sandstone datasets than on the synthetic ones. This improvement may be due to the shape factor assumptions (e.g., cuboids and pyramids) being more compatible with real sandstone grain geometries than with idealized spheres. As shown in Figures~\ref{fig:sandstone_results}a, d, g, the typical overestimation observed in the synthetic data is less pronounced. However, the PNM still exhibits moderate errors across different resolutions, as shown in Table~\ref{tab:sandstone_metrics}.

The pure data-driven GNN shows limited generalization across different pore network resolutions in the sandstone dataset. As shown in Figure~\ref{fig:sandstone_results}e, the GNN model again performs well at the training resolution ($128^3$). However, it exhibits substantial error at other resolutions (Figures~\ref{fig:sandstone_results}b and h), indicating poor generalization to unseen network sizes. As before, this is likely due to the graph-level GNN's reliance on global pooling, which can obscure topological differences between differently scaled networks.

In contrast, the GNN-embedded PNM maintains low error across all resolutions. While the model performs comparably to the GNN at the training resolution (Figures~\ref{fig:sandstone_results}c), its true advantage lies in its generalization to both smaller and larger networks (Figures~\ref{fig:sandstone_results}f,~i). This is achieved by embedding learned throat conductances into a physically consistent PNM solver, which constrains the solution based on flow physics while allowing flexibility in conductance estimation through learning. The neural network effectively replaces the empirical shape factor correlations with a data-driven mapping from pore and throat features to hydraulic conductance, improving prediction accuracy even in the absence of direct throat-scale supervision. In addition to RMSE, other evaluation metrics, including MAE, MAPE, and~$R^2$, are also reported in Table~\ref{tab:sandstone_metrics}. The results across most metrics remain consistent, further validating the accuracy and robustness of the GNN-embedded PNM model in comparison to the pure data-driven GNN and traditional PNM. One exception occurs at the coarse resolution ($64^3$), where the mean absolute percentage error (MAPE) is relatively high for the GNN-embedded PNM (93\%), despite its lower MAE and RMSE compared to the traditional PNM. The reduction in MAE and RMSE indicates that the GNN-embedded PNM mitigates large absolute errors and handles outliers more effectively. However, its MAPE is worse, which is significant because MAPE treats all samples equally by normalizing errors relative to their true permeability. Unlike RMSE, which can be dominated by a few large outliers, MAPE highlights systematic relative deviations across the entire dataset. The worse MAPE here suggests that, although the GNN-embedded PNM controls large errors well, it introduces relatively larger proportional errors for many samples when every data point is weighted equally. One likely reason is that, for such a coarse pore network, using two GNN layers may be unnecessarily deep, causing oversmoothing of node features and reducing accuracy in relative terms. In other words, the limited graph size does not require as much message-passing depth, and the extra aggregation step may dilute local structural information that is still important at this resolution. By contrast, as the resolution increases ($128^3$ and $256^3$), the GNN-embedded PNM improves consistently across all error metrics. This trend reflects that larger, more complex pore networks benefit from deeper GNNs, as additional layers help capture richer connectivity patterns and improve permeability prediction.

\begin{table}[t]
\caption{Model evaluation metrics for sandstone datasets at different resolutions using the pore network model (PNM), the pure data-driven graph neural network (GNN) model, and the GNN-embedded PNM. The metrics include mean absolute error (MAE), root mean squared error (RMSE), mean absolute percentage error (MAPE), and coefficient of determination ($R^2$).}
\label{tab:sandstone_metrics}
\centering
\begin{tabular}{p{0.15\textwidth} p{0.25\textwidth} p{0.1\textwidth} p{0.1\textwidth} p{0.1\textwidth} p{0.1\textwidth}}
\hline
\textbf{Resolution} & \textbf{Model} & \textbf{MAE} & \textbf{RMSE} & \textbf{MAPE (\%)} & \textbf{$R^2$} \\[4pt]
\hline
$64^3$ & Pore Network Model & 0.2222 & 1.2950 & 38.08 & 0.5649 \\[4pt]
   & Pure GNN & 0.3608 & 1.7828 & 686.32 & 0.1263 \\[4pt]
   & GNN-embedded PNM & 0.1259 & 0.5061 & 93.07 & 0.9296 \\[6pt]
\hline
$128^3$ &  Pore Network Model & 0.0399 & 0.1063 & 16.96 & 0.8133 \\[4pt]
    & Pure GNN & 0.0751 & 0.1142 & 172.04 & 0.7848 \\[4pt]
    & GNN-embedded PNM & 0.0383 & 0.0616 & 20.99 & 0.9373 \\[6pt]
\hline
$256^3$ &  Pore Network Model & 0.0582 & 0.1206 & 13.76 & 0.8590 \\[4pt]
    & Pure GNN & 0.0918 & 0.1445 & 36.38 & 0.7978 \\[4pt]
    & GNN-embedded PNM & 0.0537 & 0.1131 & 12.51 & 0.8761 \\[6pt]
\hline
\end{tabular}
\end{table}

Across both synthetic and real sandstone datasets, the GNN-embedded PNM consistently outperforms the pure data-driven GNN in generalization beyond the training scale. While the GNN aggregates all node and edge features into a global vector which is good for ensuring permutation invariance, it also risks losing critical multi-scale structural information. In contrast, the GNN-embedded PNM framework predicts hydraulic conductances edge-by-edge, preserving the pore network structure, and lets the PNM enforce global flow behavior via Darcy-scale mass balance. This structure-aware formulation enables better scalability across network sizes. Although the PNM component is grounded in physics, its accuracy can still suffer when empirical shape factors do not align with actual pore-throat geometries. Purely empirical corrections often fail due to the heterogeneity of real pore geometry. By replacing these empirical corrections with a GNN that learns hydraulic conductance relationships from permeability observations, the GNN-embedded PNM framework can capture complex, nonlinear dependencies—even without direct supervision at the throat level. Instead, the adjoint method provides gradients through the PNM, allowing the model to be trained solely on bulk-scale permeability data while preserving physical consistency.

\subsection{Feature Gradient Interpretation in GNN Models}

\begin{figure}
\noindent\includegraphics[width=1.0\textwidth]{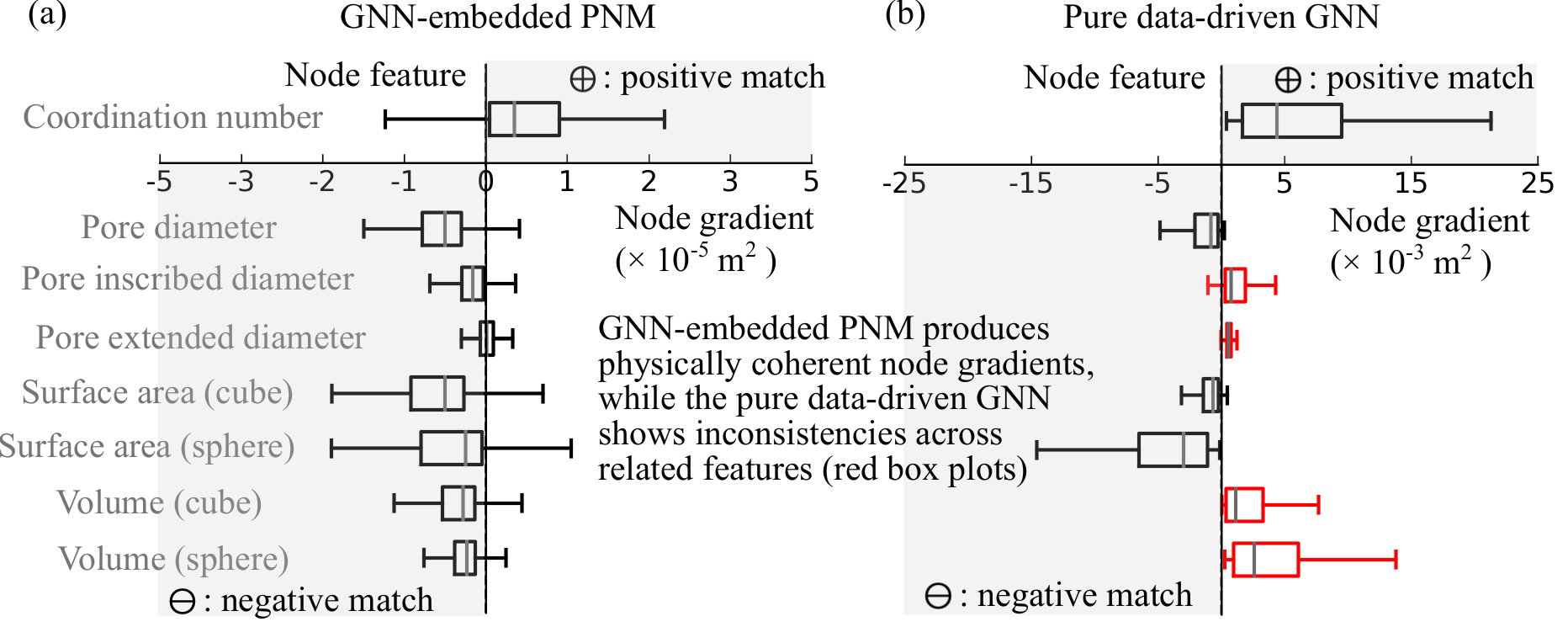}
\caption{
Permeability gradient analysis with respect to pore-level geometric features. 
(a) Results from the graph neural network (GNN)-embedded pore network model (PNM). 
(b) Results from the purely data-driven GNN model. 
Each box plot shows the distribution of~$\partial K / \partial z_i$, the partial derivative of the predicted permeability $K$ (in units of m$^2$) with respect to the z-score normalized pore feature $z_i$. 
Surface areas and volumes are computed under cube and sphere assumptions. 
The gray line inside each box represents the median value; the box edges indicate the 25th and 75th percentiles; and the whiskers extend to 1.5 times the interquartile range. 
Outliers beyond this range are omitted for clarity. 
Black box plots indicate that the gradient signs are overall consistent with physical expectations, while red box plots highlight physically inconsistent gradients across related features. 
The background shading denotes expected gradient directions: grey regions (labeled ``$\oplus$: positive match'' and ``$\ominus$: negative match'') indicate that the sign of the gradient aligns with known physical relationships---i.e., a positive gradient for a feature expected to increase permeability, or a negative gradient for one expected to reduce it.
}
\label{fig:node_gradient}
\end{figure}

\begin{figure}
\noindent\includegraphics[width=1.0\textwidth]{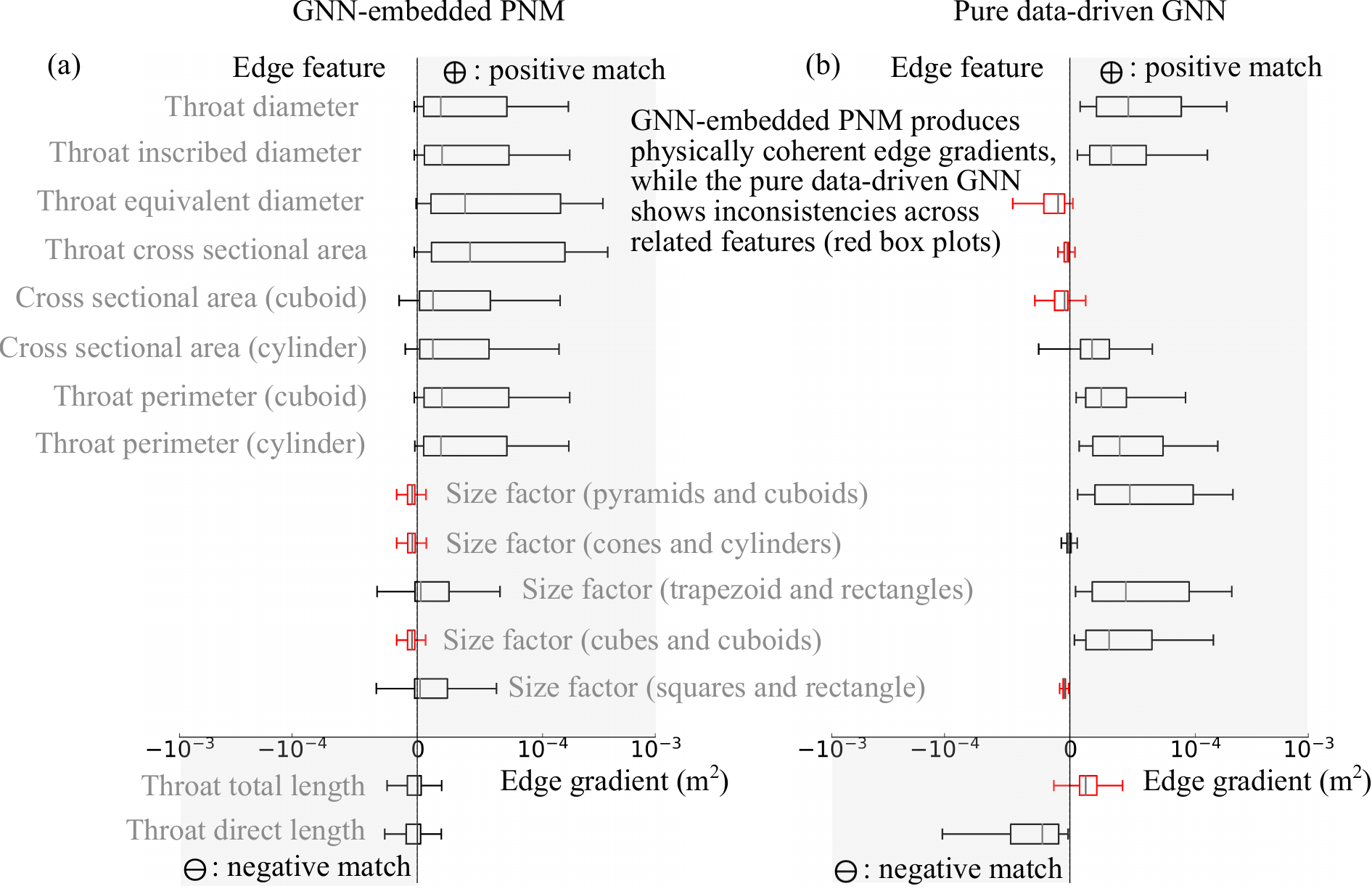}
\caption{
Permeability gradient analysis with respect to throat-level geometric features. 
(a) Results from the graph neural network (GNN)-embedded pore network model (PNM). 
(b) Results from the purely data-driven GNN model. 
Each box plot shows the distribution of~\(\partial K / \partial z_i\), the partial derivative of the predicted permeability \(K\) (in units of m\(^2\)) with respect to the z-score normalized throat feature \(z_i\). 
Cross-sectional areas are computed under general, cuboid, and cylindrical geometric assumptions; throat perimeters are computed under cuboid and cylindrical shape assumptions. 
Hydraulic size factors are derived from idealized geometric shapes, including pyramids and cuboids, cones and cylinders, trapezoids and rectangles, cubes and cuboids, and squares and rectangles. 
The gray line inside each box represents the median value; the box edges indicate the 25th and 75th percentiles; and the whiskers extend to 1.5 times the interquartile range. Outliers beyond this range are omitted for clarity. 
Black box plots indicate that the gradient signs are overall consistent with physical expectations, while red box plots highlight physically inconsistent gradients across related features. 
The background shading denotes expected gradient directions: grey regions (labeled ``$\oplus$: positive match'' and ``$\ominus$: negative match'') indicate that the sign of the gradient aligns with known physical relationships---i.e., a positive gradient for a feature expected to increase permeability, or a negative gradient for one expected to reduce it.
}

\label{fig:edge_gradient}
\end{figure}

The feature gradients show how the input features influence the output permeability for each model, providing a means to evaluate whether the predictions align with known physical relationships. To examine this, we compute and visualize the gradients of predicted permeability with respect to geometric features for both the GNN-embedded pore network model and the purely data-driven GNN. Figure~\ref{fig:node_gradient} and Figure~\ref{fig:edge_gradient} present the partial derivatives of the predicted permeability \(K\) with respect to various geometric features in the GNN-embedded pore network model and the purely data-driven graph neural network, respectively. These derivatives, \(\partial K / \partial z_i\), reflect the sensitivity of the model output to each normalized feature \(z_i\). As described in the Methods section, all node and edge features were standardized using z-score normalization in Equation~\ref{eq:zscore_normalization}; consequently, the gradient \(\partial K / \partial z_i\) has units of square meters~(m\(^2\)). In the pore network model, different pore and throat features can have either a positive or negative effect on the predicted permeability \(K\). To visualize how the model predictions align with these physical expectations, we separate features based on their anticipated influence: features expected to increase permeability are plotted on the positive \(y\)-axis, while those expected to decrease it are plotted on the negative \(y\)-axis. In the figure, grey regions indicate a model prediction that aligns with a expected effect, while no-color regions indicate a prediction inconsistent with a expected effect. The box plots summarize the distribution of gradient values across all samples for each feature. The central grey line in each box indicates the median value, while the edges of the box represent the 25th (lower quartile) and 75th (upper quartile) percentiles. The whiskers extend to 1.5 times the interquartile range (IQR) from the lower and upper quartiles, capturing the typical range of values. Outliers beyond this range have been excluded to enhance visibility. Box plots shown in red highlight features where the model’s predicted gradient direction is inconsistent with physical expectations. This visualization provides a comparative overview of how each node and edge feature contributes to permeability prediction across different samples and how well the models conform to known physical trends.

The GNN-embedded pore network model captures physically meaningful and interpretable sensitivities to pore-scale geometric features. Figure~\ref{fig:node_gradient}a presents the gradients of predicted permeability with respect to various pore features, as computed by the GNN-embedded pore network model. The results indicate that pore diameter, surface area, and volume features have median gradient values that are negative. This suggests that, within the learned GNN-embedded pore network model, increasing the pore diameter tends to decrease the predicted global permeability \(K\). One possible explanation is that a larger pore may correspond to a longer effective flow path, which reduces local hydraulic conductance and thereby lowers overall permeability. Since surface area and pore volume are both functions of pore diameter, they exhibit similar trends. In contrast, the gradient with respect to coordination number exhibits a positive median value in Figure~\ref{fig:node_gradient}a. Physically, a higher coordination number reflects more neighboring connections per pore, which improves flow connectivity and reduces resistance. As a result, increasing the coordination number raises the predicted permeability \(K\), consistent with theoretical expectations from pore network transport models. Overall, the gradients in Figure~\ref{fig:node_gradient}a exhibit median values that are consistent with physical intuition. Each feature's median value falls within the shaded region that matches its expected directional effect, indicating that the model has learned interpretable, physically meaningful sensitivities.

The purely data-driven GNN exhibits inconsistent and physically implausible node gradient behavior, reflecting its lack of built-in physical constraints. Figure~\ref{fig:node_gradient}b presents the gradient analysis for the purely data-driven GNN model, which does not incorporate a flow solver. In this model, the partial derivatives with respect to the three diameter-related features are inconsistent: the gradient with respect to pore diameter has a negative median value, while those with respect to pore inscribed diameter and pore extended diameter have positive medians. Similarly, pore surface area and pore volume, which are both functions of pore diameter, exhibit contradictory behavior: the gradient with respect to pore volume is positive, whereas that for surface area is negative. These inconsistencies are difficult to interpret from a physical perspective. Such conflicting patterns suggest that, in the absence of explicit physical constraints, the purely data-driven GNN may learn feature relationships that do not conform to fluid-mechanical principles. As a result, the model tends to fit statistical correlations present in the training data, rather than capturing physically meaningful behavior. Compared to the GNN-embedded pore network model, the gradients in the pure GNN model are concentrated in a smaller subset of features, such as coordination number and the volume and surface area under spherical assumptions. This suggests that the model tends to treat graph-level permeability prediction as a global pooling operation over node and edge features, often relying heavily on the single most predictive parameter. In contrast, the GNN-embedded model exhibits a more balanced gradient distribution across all features, indicating a broader and more physically grounded use of the input information. The more balanced importance across features makes the GNN-embedded pore network model better suited for extrapolation to conditions or geometries that are not well represented in the training set.

% To understand how throat-level geometric features influence predicted permeability, we analyze the sensitivity of the model output with respect to various throat attributes, as shown in Figure~\ref{fig:edge_gradient}. These features include throat diameter, inscribed diameter, and equivalent diameter; cross-sectional areas computed under general, cuboid, and cylindrical geometric assumptions; throat perimeters derived from cuboid and cylindrical shapes; and hydraulic size factors based on idealized geometries, such as pyramids and cuboids, cones and cylinders, trapezoids and rectangles, cubes and cuboids, and squares and rectangles. Additionally, we consider both the throat total length and the throat direct length.

The GNN-embedded pore network model exhibits physically consistent gradient behavior across most throat-level geometric features. Figure~\ref{fig:edge_gradient}a corresponds to this model. Throat diameters and cross sectional areas generally exhibit positive gradient values, which aligns with the understanding that enlarging a conduit or increasing its cross sectional area reduces flow resistance, thereby enhancing permeability. Throat perimeter often follows a similar trend. An increase in perimeter can improve local hydraulic conductance~\(g\) as the perimeter associated with an expanded cross section. In contrast, features related to throat length tend to show negative gradients. As the fluid must travel farther through longer throats, the associated pressure drop increases, resulting in reduced overall permeability. The hydraulic size factors gradients show both positive and negative signs. The hydraulic size factor $F$ is the geometric part of the hydraulic conductance, shown in Equation~\ref{eq:hydraulic_size_factor}. Ideally, permeability increases with hydraulic size factor. As they are defined by idealized geometries, the results from the GNN-embedded pore network model indicate these geometry may not represent the real pore and throat shapes very well. Nevertheless, if needed, we can still enforce monotonic constraints so that larger size factors always increase hydraulic conductance. This flexibility arises from the explicit presence of hydraulic conductance $g$ in our pipeline: we can make the gradient of hydraulic conductance with respect to the feature~$\frac{\partial g}{\partial \mathrm{F}} \geq 0$ simply by adopting a monotonic neural-network architecture or by imposing an appropriate regularization term. In a PDE-free setup, there is no direct handle on $g$ to constrain in that manner. Compared with hydraulic size factor, the largest gradients appear on throat diameter, cross sectional area, and throat perimeter, indicating that the GNN sees these as the most critical features for controlling flow.

The purely data-driven GNN model produces edge gradient patterns that are often physically inconsistent. Figure~\ref{fig:edge_gradient}b shows the edge gradient analysis for the purely data-driven GNN model, which directly maps input features to the global permeability \(K\) without incorporating a flow solver. In the absence of physical constraints, the resulting partial derivatives can become inconsistent across features that are physically related and would normally exhibit similar behavior. For instance, the gradients of permeability with respect to throat diameter and throat inscribed diameter are positive, while the gradient with respect to throat equivalent diameter is negative, despite all three representing similar geometric properties. A similar inconsistency is observed in the cross-sectional area features. The gradients associated with general and cuboid-based cross-sectional areas are predominantly negative, even though larger cross-sectional areas typically enhance hydraulic conductance by widening the flow path. In contrast, the gradient with respect to the cylinder-based cross-sectional area is positive, which is more physically plausible. These contradictions also appear in the length-related features. Throat length is expected to exhibit a negative gradient due to its role in increasing flow resistance. However, throat direct length in pure GNN model case unexpectedly shows a positive gradient. These inconsistencies suggest that the pure GNN model is primarily capturing statistical correlations from the training data rather than physically meaningful relationships. The model tends to assign large positive or negative weights to whichever features are most correlated with permeability in the data, even if those relationships contradict physical principles. Moreover, the magnitudes of gradients across similar types of features vary more significantly in the pure data-driven GNN model than in the GNN-embedded counterpart. For example, the gradient magnitude for throat equivalent diameter is much smaller than that for throat diameter or inscribed diameter. Similarly, the gradient range for throat direct length is much larger than that for throat total length. In contrast, the GNN-embedded pore network model produces more consistent gradient magnitudes across related features, as shown in Figure~\ref{fig:edge_gradient}a. This consistency indicates that the GNN-embedded pore network model yields more physically coherent predictions, which is particularly advantageous when applying the model to unseen structures governed by the same underlying physical processes.
\section{Conclusions}
In this work, we proposed a graph neural network-embedded pore network model for predicting the permeability of porous media. Unlike purely data-driven approaches or traditional physics-based models, this hybrid framework combines the flexibility of machine learning with the physical interpretability of a solver. The GNN is used to learn the complex relationship between pore geometry and hydraulic conductance, replacing the conventional analytical expressions based on simplified geometric assumptions. These predicted hydraulic conductance values are then used as inputs to the PNM solver, which computes the overall permeability.

This hybrid model addresses key limitations faced by both standalone GNNs and classical PNMs. Pure data-driven graph neural network models tend to suffer from accuracy degradation when applied to datasets with different spatial dimensions, as they collapse rich graph-level features into a single scalar output, which can result in loss of structural information. In contrast, the GNN-embedded pore network model preserves the pore network’s graph structure from input to output, which improves generalization performance on unseen datasets and maintains higher prediction accuracy across varying dimensions. On the other hand, classical PNMs, while computationally efficient and physically grounded, rely on oversimplified pore geometry assumptions, which can lead to significant errors in hydraulic conductance estimation. By learning hydraulic conductance values directly from data, our method enhances the PNM’s ability to capture the complex relationship between pore geometry and conductance, avoiding the use of simplified geometric assumptions that often introduce model uncertainty. The model is trained end-to-end using only permeability as supervision. This is achieved by computing the gradient of the loss function with respect to the GNN parameters through the application of the chain rule. Specifically, we use a discrete adjoint method to compute gradients of the permeability output with respect to the predicted hydraulic conductance values. The gradients are then backpropagated through the GNN via automatic differentiation. This training process eliminates the need for computationally expensive, pore-level hydraulic conductance simulations, improving scalability and practicality for large datasets. Furthermore, sensitivity analysis demonstrates that the GNN-embedded pore network model yields more consistent and physically meaningful responses to different pore and throat features compared to the pure data-driven GNN model. This reinforces the reliability of the hybrid framework and highlights the advantage of incorporating physical solvers into data-driven modeling pipelines—not only to improve predictive performance but also to enhance model interpretability and robustness.

Overall, this study highlights the potential of hybrid physics–machine learning approaches for modeling flow in porous media. By embedding neural networks within physically grounded solvers, we can develop models that are accurate, generalizable, and physically interpretable. This framework opens new opportunities for model closure, uncertainty reduction, and scalable learning in complex subsurface flow systems.

\section*{Data Availability Statement}
The graph training dataset and the corresponding ground-truth permeability values used in this work are publicly available via Zenodo: https://doi.org/10.5281/zenodo.16610986.
The GitHub repository (https://github.com/ITLR-DDSim/gnn-pnm-permeability) contains the full source code, training scripts, and documentation required to reproduce and extend the GNN-based permeability prediction framework.

\acknowledgments
This work is funded by Deutsche Forschungsgemeinschaft (DFG, German Research Foundation) under Germany’s Excellence Strategy- EXC 2075– 390740016. We acknowledge the support by the Stuttgart Center for Simulation Science (SimTech).

\appendix
\section{Pore and Throat Features}
\label{app:features}

Appendix A provides detailed descriptions of the properties used as node and edge features in the graph representation of porous media, utilized for GNN training. Table \ref{tab:pore-properties} summarizes the pore properties assigned to nodes, while Table \ref{tab:throat-properties} outlines the throat properties assigned to edges. Throat hydraulic size factors and other pore and throat properties were computed based on various shape factor assumptions to capture the geometric diversity within the porous structure.

\begin{table}[t]
\caption{Description of pore properties used in deep learning model training.}
\label{tab:pore-properties}
\centering
\begin{tabular}{p{0.30\textwidth} p{0.65\textwidth}}
\hline
\textbf{Property name} & \textbf{Description} \\
\hline
Pore diameter & Equivalent diameter of the pore. \\
Pore inscribed diameter & Diameter of the largest sphere fully inscribed within the pore. \\
Pore extended diameter & Diameter of the largest sphere encompassing the pore region. \\
Coordination number & Number of throats connected to the pore. \\
Surface area (cube shape factor) &  Internal surface area of pore bodies calculated assuming the pore shape approximates a cube. \\
Surface area (sphere shape factor) &  Internal surface area of pore bodies calculated assuming the pore shape approximates a sphere. \\
Volume (cube shape factor) & Space or void within the pore calculated assuming the pore shape approximates a cube. \\
Volume (sphere shape factor) & Space or void within the pore calculated assuming the pore shape approximates a sphere. \\
\hline
\end{tabular}
\end{table}

\begin{table}[t]
\caption{Description of throat properties used in the deep learning model training.}
\label{tab:throat-properties}
\centering
\begin{tabular}{p{0.30\textwidth} p{0.65\textwidth}}
\hline
\textbf{Property name} & \textbf{Description} \\
\hline
Throat diameter & Equivalent diameter of the throat. \\
Throat inscribed diameter & Diameter of the largest sphere fully inscribed within the throat. \\
Throat total length & Total length of the throat path between connected pores. \\
Throat direct length & Direct Euclidean distance between connected pores. \\
Throat cross-sectional area & Cross-sectional area of the throat opening. \\
Throat equivalent diameter & Equivalent diameter derived from cross-sectional area. \\
Hydraulic size factor (pyramids and cuboids) & Hydraulic size factor calculated assuming throat geometry as pyramids and cuboids. \\
Hydraulic size factor (cones and cylinders) & Hydraulic size factor calculated assuming throat geometry as cones and cylinders. \\
Hydraulic size factor (trapezoids and rectangles) & Hydraulic size factor calculated assuming throat geometry as trapezoids and rectangles. \\
Hydraulic size factor (cubes and cuboids) & Hydraulic size factor calculated assuming throat geometry as cubes and cuboids. \\
Hydraulic size factor (squares and rectangles) & Hydraulic size factor calculated assuming throat geometry as squares and rectangles. \\
Throat area (cuboid shape factor) & Throat cross-sectional area assuming a cuboid shape. \\
Throat area (cylinder shape factor) & Throat cross-sectional area assuming a cylindrical shape. \\
Throat perimeter (cuboid shape factor) & Boundary length of the throat calculated assuming throat cross-section is cuboidal. \\
Throat perimeter (cylinder shape factor) & Boundary length of the throat calculated assuming throat cross-section is cylindrical. \\
\hline
\end{tabular}
\end{table}

\section{Detailed Graph Neural Network Architecture}
\label{app:gnn}
\begin{figure}
\noindent\includegraphics[width=\textwidth]{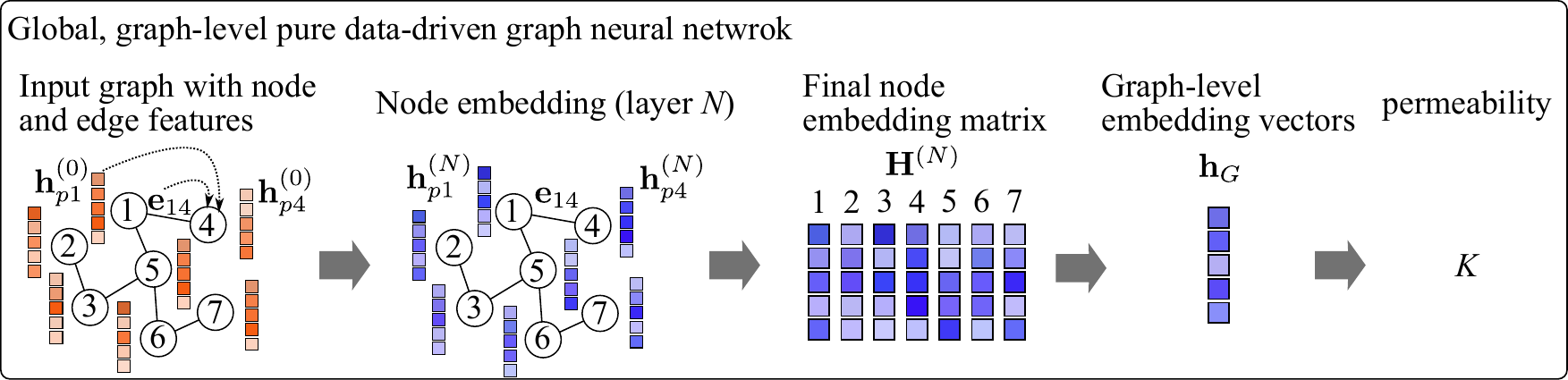}
\caption{
Illustration of the workflow for a global, graph-level pure data-driven graph neural network. In the graph-level graph neural network model, the input graph with initial node features $\mathbf{h}^{(0)}$ and edge features $\mathbf{e}$ is processed through $N$ graph neural network layers to obtain final node embeddings $\mathbf{h}^{(N)}$. These embeddings are aggregated into a node embedding matrix~$\mathbf{H}^{(N)}$, then pooled to a graph-level embedding vector $\mathbf{h}_{G}$, which is used to predict the permeability scalar $K$. 
}
\label{fig:gnn1}
\end{figure}

In this Appendix, we provide details of the two GNN models we used in this study for the permeability prediction. Figure~\ref{fig:gnn1} shows the workflow of the global, graph-level pure data-driven GNN. 
In the pure data-driven GNN model, the same message-passing steps as in the GNN embedded-pore network model (Equations~\ref{eq:node_message} and~\ref{eq:node_update}) are employed. After \( N \) GNN layers, we obtain the final node embeddings \(\mathbf{h}_p^{(N)}\), which are collected into a node embedding matrix \(\mathbf{H}^{(N)}\). A readout function then pools these embeddings to produce a graph-level representation, which is used to predict the permeability scalar \( K \):
\begin{linenomath*}
\begin{equation}
K = \sigma\left[\beta + \mathbf{w}_K^\top \left(\tfrac{1}{P}\, \mathbf{H}^{(N)}\, \mathbf{1}\right)\right],
\label{eq:graph_invariance}
\end{equation}
\end{linenomath*}
where \( \mathbf{1} \) is an all-ones vector, \( P \) is the number of nodes in the graph, and \( \beta \), \( \boldsymbol{\mathbf{w}}_K \), and~\( \sigma \) are learnable parameters. The graph-level GNN algorithm is given in Table~\ref{tab:gnn-architecture}.

\begin{table}[t]
\caption{Two-layer global, graph-level pure data-driven graph neural network architecture for permeability prediction.  
Each node embedding layer updates \(\mathbf{h}_p^{(\ell)}\), and a final readout layer aggregates the node embeddings into a graph-level feature vector for predicting the permeability scalar \(K\). The ``msg'' step aggregates messages from neighboring nodes and edges, and the ``apply'' step updates each node's embedding.}
\label{tab:gnn-architecture}
\centering
\begin{tabular}{p{0.30\textwidth} p{0.45\textwidth} p{0.20\textwidth}}
\hline
\textbf{Layer Type} & \textbf{Function} & \textbf{Dimensions} \\
\hline
\textbf{Node Embedding: Layer 1} & & \\[-2pt]
msg(1)
& $\displaystyle \mathbf{H}_p^{(1)} = f_{\mathrm{msg,1}}\bigl([\mathbf{h}_{p'}^{(0)};\,\mathbf{e}_{pp'}],\,\mathbf{w}_{\mathrm{msg,1}}\bigr)$ 
& $(d_h^{(0)} + d_e)\to d_h^{(0)}$\\
Activation
& ReLU 
& $-$ \\[4pt]
apply(1) 
& $\displaystyle \mathbf{h}_p^{(1)} = f_{\mathrm{apply,1}}\bigl([\mathbf{h}_p^{(0)};\,\mathbf{H}_p^{(1)}],\,\mathbf{w}_{\mathrm{apply,1}}\bigr)$
& $2\,d_h^{(0)} \to d_h^{(1)}$\\
Activation
& ReLU 
& $-$ \\[6pt]

\textbf{Node Embedding: Layer 2} & & \\[-2pt]
msg(2)
& $\displaystyle \mathbf{H}_p^{(2)} = f_{\mathrm{msg,2}}\bigl([\mathbf{h}_{p'}^{(1)};\,\mathbf{e}_{pp'}],\,\mathbf{w}_{\mathrm{msg,2}}\bigr)$ 
& $(d_h^{(1)} + d_e)\to d_h^{(1)}$\\
Activation
& ReLU 
& $-$ \\[4pt]
apply(2)
& $\displaystyle \mathbf{h}_p^{(2)} = f_{\mathrm{apply,2}}\bigl([\mathbf{h}_p^{(1)};\,\mathbf{H}_p^{(2)}],\,\mathbf{w}_{\mathrm{apply,2}}\bigr)$ 
& $2\,d_h^{(1)} \to d_h^{(2)}$\\
Activation
& ReLU
& $-$ \\[6pt]

\textbf{Readout: Graph-Level Prediction} & & \\[-2pt]
Pooling
& $\displaystyle \mathbf{h}_G = \tfrac{1}{P} \mathbf{H}^{(2)}\, \mathbf{1}$ 
& $P \times d_h^{(2)} \to d_h^{(2)}$\\
MLP (Output)
& $\displaystyle K = \sigma\left[\beta + \boldsymbol{\mathbf{w}}_K^\top \mathbf{h}_G\right]$
& $d_h^{(2)} \to 1$\\
Activation
& Softplus
& $-$ \\
\hline
\end{tabular}
\end{table}

Compared to CNNs, graph neural networks naturally support stronger symmetry properties for graph-based inputs.  
In a graph neural network, each node aggregates messages from its neighbors using a shared function \(f_{\mathrm{msg}}\) in Equation~\ref{eq:node_message}, which is applied identically to all connected edges.  
This shared update mechanism means a node's embedding depends only on its neighbors' features, which makes the layer permutation equivariant.  
The aggregated messages are then combined using a symmetric function such as sum, mean, or max. Therefore, the aggregation result is unchanged regardless of the order in which neighbors are listed, ensuring permutation invariance.  
It means that the only relationship considered between two nodes is whether they are connected, not their angle, orientation, or location in space.  
As a result, GNNs exhibit effective rotation invariance for pore network graphs, since node orientation is never encoded.  
By contrast, CNNs use kernels with position-specific weights applied to structured grids.  
Each pixel’s contribution depends on its relative position in the kernel window, so rotating or permuting an image alters the result.  
CNNs therefore require data augmentation (e.g., flipping or rotation) to enforce approximate invariance, while GNNs inherently achieve this through their architecture.

However, this invariance to node order comes at a cost in graph-level prediction tasks.  
While the relative structure is implicitly encoded in the connectivity graph, once the node embeddings are collected into the final matrix \(\mathbf{H}^{(N)}\) in Figure~\ref{fig:gnn1}, no explicit spatial or structural information remains.  
The node indices in \(\mathbf{H}^{(N)}\) are arbitrary and do not reflect any position or topological context. Therefore, graph-level GNNs are well suited for classification tasks where such invariance is desirable, but may be less effective for regression problems like permeability prediction, which often depend on fine-grained positional and structural information. 
In contrast, CNNs preserve spatial layout from start to finish. Even after several convolution and pooling layers, the final fully connected layer receives a flattened image in a fixed spatial order, so it can still exploit coarse positional patterns such as top-versus-bottom or left-versus-right. 
Thus, although GNNs offer permutation and rotation invariance to reduce the need for data augmentation, they lose access to absolute node positional and structural information when making graph-level predictions.

The GNN-embedded pore network model is described in Section~\ref{sec:gnn}. For a more detailed overview of its architecture, the edge-level GNN algorithm is further outlined in Table~\ref{tab:gnn-architecture2}.

\begin{table}[t]
\caption{Two-layer local, edge-level graph neural network architecture for predicting edge-wise conductances $g_i$.  
Each node embedding layer updates $\mathbf{h}_p^{(k)}$, and the final edge predictor outputs a scalar per throat. The ``msg'' step computes messages from neighboring nodes and edges, while the ``apply'' step updates each node’s embedding. After two such layers, a two-step MLP generates the final edge-level scalar conductance.}
\label{tab:gnn-architecture2}
\centering
\begin{tabular}{p{0.30\textwidth} p{0.40\textwidth} p{0.25\textwidth}}
\hline
\textbf{Layer Type} & \textbf{Function} & \textbf{Dimensions} \\
\hline
\textbf{Node Embedding: Layer 1} & & \\[-2pt]
msg(1)
& $\displaystyle \mathbf{H}'_p = f_{\mathrm{msg,1}}\!\bigl([\mathbf{h}_{p'}^{(0)};\,\mathbf{e}_{pp'}],\,\mathbf{w}_{\mathrm{msg,1}}\bigr)$ 
& $(d_{h}^{(0)} + d_e)\to d_{h}^{(0)}$\\
Activation
& ReLU 
& $-$ \\[4pt]
apply(1) 
& $\displaystyle \mathbf{h}_p^{(1)} = f_{\mathrm{apply,1}}\!\bigl([\mathbf{h}_p^{(0)};\,\mathbf{H}'_p],\,\mathbf{w}_{\mathrm{apply,1}}\bigr)$
& $(2\,d_{h}^{(0)})\to d_{h}^{(1)}$\\
Activation
& ReLU 
& $-$ \\[6pt]

\textbf{Node Embedding: Layer 2} & & \\[-2pt]
msg(2)
& $\displaystyle \mathbf{H}'_p = f_{\mathrm{msg,2}}\!\bigl([\mathbf{h}_{p'}^{(1)};\,\mathbf{e}_{pp'}],\,\mathbf{w}_{\mathrm{msg,2}}\bigr)$ 
& $(d_{h}^{(1)} + d_e)\to d_{h}^{(1)}$\\
Activation
& ReLU 
& $-$ \\[4pt]
apply(2)
& $\displaystyle \mathbf{h}_p^{(2)} = f_{\mathrm{apply,2}}\!\bigl([\mathbf{h}_p^{(1)};\,\mathbf{H}'_p],\,\mathbf{w}_{\mathrm{apply,2}}\bigr)$ 
& $(2\,d_{h}^{(1)})\to d_{h}^{(2)}$\\
Activation
& ReLU
& $-$ \\[6pt]

\textbf{Edge Predictor} & & \\[-2pt]
MLP (Hidden)
& $\displaystyle \mathbf{m}_{pp'} = f_{\mathrm{hid}}\!\bigl([\mathbf{h}_{p}^{(2)};\,\mathbf{h}_{p'}^{(2)};\,\mathbf{e}_{pp'}],\,\mathbf{w}_{\mathrm{hid}}\bigr)$
& $(2\,d_{h}^{(2)} + d_e)\to d_p$\\
Activation
& ReLU
& $-$ \\[4pt]
MLP (Output)
& $\displaystyle g_{pp'} = f_{\mathrm{out}}\!\bigl(\mathbf{m}_{pp'},\,\mathbf{w}_{\mathrm{out}}\bigr)$
& $d_p\to 1$\\
Activation
& Softplus
& $-$ \\
\hline
\end{tabular}
\end{table}

% In this approach, the GNN predicts edge-wise hydraulic conductances using the final node embeddings and edge features. These predicted hydraulic conductances are then used as inputs to a pore network model solver, which computes the resulting permeability. By preserving the full pore network structure and explicitly modeling the physical flow mechanism through the pore network model, this hybrid approach enables more structure-aware permeability predictions. 

\section{Detailed Pore-Network Model Formulation and Permeability Computation}
\label{app:PNM}
In this appendix, we provide a detailed derivation of the linear system used in the pore network model and describe how the overall permeability is computed once the pore pressures are known. Consider a network of \(N_p\) pores, each labeled by an index \(p\), and~\(N_t\) throats, each labeled by \(i\). Let \(\boldsymbol{\Gamma}_{\mathrm{inlet}}\) and \(\boldsymbol{\Gamma}_{\mathrm{outlet}}\) be the sets of inlet and outlet pores, and let \(\boldsymbol{\Gamma}\) represent all boundary pores, which include both inlets and outlets. For any internal pore \(p\notin\boldsymbol{\Gamma}\), mass conservation requires that the net inflow to that pore is zero. If \(\mathbf{T}_p\) is the set of throats incident on pore \(p\), and \(q_i\) is the flow rate through throat \(i\), the pore-level mass-balance condition is
\begin{linenomath*}
\begin{equation}
\sum_{i \in \mathbf{T}_p} q_i = 0 \label{eq:mass_conservation}
\end{equation}
\end{linenomath*}
Each throat \(i\) connects two pores, say \(p_1\) and \(p_2\). Under a linear (Darcy-type) flow assumption, the flow rate through throat \(i\) is given by
\begin{linenomath*}
\begin{equation}
q_i = g_i \bigl(x_{p_1} - x_{p_2}\bigr) \label{eq:flux_definition}
\end{equation}
\end{linenomath*}
where \(x_p\) is the pressure at pore \(p\) and \(g_i\) is the constant conductance of throat \(i\). The boundary conditions specify pressures on the inlet and outlet pores:
\begin{linenomath*}
\begin{equation}
x_p = P_{\mathrm{inlet}} \quad \forall\, p \in \boldsymbol{\Gamma}_{\mathrm{inlet}}, \quad
x_p = P_{\mathrm{outlet}} \quad \forall\, p \in \boldsymbol{\Gamma}_{\mathrm{outlet}} \label{eq:bc_pressure}
\end{equation}
\end{linenomath*}
where $P_{\mathrm{inlet}}$ is inlet pressure and $P_{\mathrm{outlet}}$ is outlet pressure. By collecting the mass-conservation equations for all internal pores and enforcing these boundary pressures on the inlet and outlet pores, we assemble the matrix \(\mathbf{A}\) and vector \(\mathbf{b}\) so that they reflect the conductance contributions from every throat and the prescribed boundary conditions on \(\boldsymbol{\Gamma}_{\mathrm{inlet}}\) and~\(\boldsymbol{\Gamma}_{\mathrm{outlet}}\). This results in the following linear system:
\begin{linenomath*}
\begin{equation}
\mathbf{A}\,\mathbf{x} = \mathbf{b} \label{eq:Ax_equal_b}
\end{equation}
\end{linenomath*}
Once this linear system is formed, solving it yields the pore-pressure vector \(\mathbf{x}\). Then, the total flow rate \(Q_{\mathrm{in}}\) through the inlet is computed by summing all throats that connect to the inlet pores:
\begin{linenomath*}
\begin{equation}
Q_{\mathrm{in}} = \sum_{i \in \mathbf{T}_{\mathrm{inlet}}} g_i \bigl(x_{p_1} - x_{p_2}\bigr) \label{eq:Qin}
\end{equation}
\end{linenomath*}
where \(\mathbf{T}_{\mathrm{inlet}}\) is the set of throats touching the inlet pores, and \(p_1\) and \(p_2\) are the pores connected by each throat \(i\). Under steady-state conditions, \(Q_{\mathrm{in}} = Q_{\mathrm{out}}\), and we denote this common flow rate by \(Q\). At the macroscale, Darcy's law states 
\begin{linenomath*}
\begin{equation}
Q = \frac{K\,A_s}{\mu\,L}\,\Delta P \label{eq:darcy_law}
\end{equation}
\end{linenomath*}
where \(K\) is the permeability, \(A_s\) is the cross-sectional area, \(\mu\) is the fluid viscosity, \(L\) is the length of the domain in the primary flow direction, and \(\Delta P = x_{\mathrm{inlet}} - x_{\mathrm{outlet}}\) is the pressure drop. Introducing the a parameter $c = \frac{\mu\,L}{A_s \Delta P}$, Darcy's law simplifies to: 
\begin{linenomath*}
\begin{equation}
Q = \frac{K}{c} \label{eq:Q_cK}
\end{equation}
\end{linenomath*}
Solving for \(K\) gives
\begin{linenomath*}
\begin{equation}
K = cQ \label{eq:K_from_Q}
\end{equation}
\end{linenomath*}
Thus, once the pore pressures are obtained from the linear system \(\mathbf{A}\,\mathbf{x} = \mathbf{b}\), the flow rate \(Q_{\mathrm{in}}\) (equal to \(Q\) in steady state) is readily evaluated and substituted into the above formula to find the overall permeability \(K\).

\section{Derivation of Discrete Adjoint System for Pore Network Model}
\label{app:adjoint}

The objective of this derivation is to compute the sensitivity of the permeability loss function \( J = \|K - K^*\|^2 \) with respect to the hydraulic conductances \( g_i \), i.e., \( \frac{dJ}{d\mathbf{g}} \). This gradient is essential for training the graph neural network, as it enables the conductances \( g_i \) to be updated in a manner that aligns the predicted permeability \(K\) with the target permeability \(K^*\). To obtain this sensitivity efficiently, we employ the discrete adjoint method.

The derivation of the discrete adjoint method from an analytical forward problem can follow one of two possible pathways. In the first approach, one derives the adjoint equations from the continuous PDE form of the forward model, and then applies a discretization scheme to both the forward and adjoint systems. In the second approach, one discretizes the forward problem (e.g., by finite volumes or pore network equations) first and then derives the corresponding discrete adjoint equations. Whenever the forward problem is already discretized for a numerical solution, obtaining the discrete adjoint equations is straightforward, as will be demonstrated. By contrast, deriving a continuous adjoint often requires extensive analytical work for each update of the forward model and involves careful derivations of boundary conditions. Therefore, this work focuses on deriving the adjoint equations based on the discretized forward problem and does not further discuss the continuous adjoint approach.

From~\ref{app:PNM}, the PNM forward model is written as a linear system
\begin{linenomath*}
\begin{equation}
\mathbf{r}(\mathbf{x}, \mathbf{g}) = \mathbf{A}(\mathbf{g})\mathbf{x} - \mathbf{b} = 0 \label{eq:residual}
\end{equation}
\end{linenomath*}
where $\mathbf{x}$ is the state vector (e.g., a vector of pressures), and hydraulic conductivity $\mathbf{g}$ is a parameter vector upon which the matrix $\mathbf{A}(\mathbf{g})$ depends. Additionally,  the PNM permeability $K$ depends on the state vector $\mathbf{x}$. Finally, The loss function to be minimized is then written as $J(K(\mathbf{x}), \mathbf{g}).$ A typical inverse or data assimilation problem (estimating hydraulic conductivity $\mathbf{g}$ from observations) is posed as the following constrained optimization:

\begin{linenomath*}
\begin{equation}
\begin{array}{ll}
\displaystyle\mathop{\min}_{\mathbf{g}} & J(K(\mathbf{x}), \mathbf{g}) \\
\textrm{subject to} & \mathbf{r}(\mathbf{x}, \mathbf{g}) = 0
\end{array} \label{eq:optimization_problem}
\end{equation}
\end{linenomath*}
Our aim is to compute the gradient of loss $J$ with respect to hydraulic conductivity $\mathbf{g}$ to facilitate iterative optimization updates for $\mathbf{g}$ within the GNN-PNM training framework. To enforce the constraint~$\mathbf{r}(\mathbf{x},\mathbf{g})=0$ when differentiating loss $J$, we introduce a Lagrange multiplier (adjoint variable) $\boldsymbol{\lambda}$ and define the Lagrangian:

\begin{linenomath*}
\begin{equation}
\mathcal{L}(\mathbf{x}, \mathbf{g}, \boldsymbol{\lambda}) = J(K(\mathbf{x}), \mathbf{g}) - \boldsymbol{\lambda}^T \mathbf{r}(\mathbf{x}, \mathbf{g}) \label{eq:lagrangian}
\end{equation}
\end{linenomath*}
At the solution, $\mathbf{r}(\mathbf{x},\mathbf{g})=0$, thus the gradient of loss $J$ with respect to hydraulic conductivity $\mathbf{g}$ can be replaced by that of Lagrangian $\mathcal{L}$:

\begin{linenomath*}
\begin{equation}
\frac{dJ}{d\mathbf{g}} = \frac{d\mathcal{L}}{d\mathbf{g}} \label{eq:dJ_dg}
\end{equation}
\end{linenomath*}
Hence, we differentiate Lagrangian $\mathcal{L}$ with respect to each parameter component hydraulic conductivity $g_i$:

\begin{linenomath*}
\begin{equation}
\frac{dJ}{dg_i} = \frac{\partial J}{\partial g_i} + \frac{\partial J}{\partial \mathbf{x}}\frac{d\mathbf{x}}{dg_i}  
- \boldsymbol{\lambda}^T \frac{d\mathbf{r}}{dg_i} 
- \frac{d\boldsymbol{\lambda}^T}{dg_i} \mathbf{r} \label{eq:total_dJ_dgi}
\end{equation}
\end{linenomath*}
Since constraint $r(\mathbf{x},\mathbf{g})=0$ at the solution, the term involving $\frac{d\boldsymbol{\lambda}^T}{dg_i}\mathbf{r}$ vanishes. Expanding $\frac{d\mathbf{r}}{dg_i}$, we have

\begin{linenomath*}
\begin{equation}
\frac{d\mathbf{r}}{dg_i} = \frac{\partial \mathbf{r}}{\partial g_i} + \frac{\partial \mathbf{r}}{\partial \mathbf{x}}\frac{d\mathbf{x}}{dg_i} \label{eq:total_dr_dgi}
\end{equation}
\end{linenomath*}
thus,

\begin{linenomath*}
\begin{equation}
\frac{dJ}{dg_i} = \frac{\partial J}{\partial g_i} + \frac{\partial J}{\partial \mathbf{x}}\frac{d\mathbf{x}}{dg_i} 
- \boldsymbol{\lambda}^T\left(\frac{\partial \mathbf{r}}{\partial g_i} + \frac{\partial \mathbf{r}}{\partial \mathbf{x}}\frac{d\mathbf{x}}{dg_i}\right) \label{eq:total_dJ_dgi_compact}
\end{equation}
\end{linenomath*}
Next, we group terms involving $\frac{d\mathbf{x}}{dg_i}$:

\begin{linenomath*}
\begin{equation}
\frac{dJ}{dg_i} = \frac{\partial J}{\partial g_i} - \boldsymbol{\lambda}^T \frac{\partial \mathbf{r}}{\partial g_i} 
+ \left(\frac{\partial J}{\partial \mathbf{x}} - \boldsymbol{\lambda}^T \frac{\partial \mathbf{r}}{\partial \mathbf{x}}\right) \frac{d\mathbf{x}}{dg_i} \label{eq:adjoint_simplification}
\end{equation}
\end{linenomath*}
We now choose Lagrange multiplier $\boldsymbol{\lambda}$ to eliminate the $\frac{d\mathbf{x}}{dg_i}$ term by setting:

\begin{linenomath*}
\begin{equation}
\frac{\partial J}{\partial \mathbf{x}} - \boldsymbol{\lambda}^T \frac{\partial \mathbf{r}}{\partial \mathbf{x}} = 0 \label{eq:adjoint_condition}
\end{equation}
\end{linenomath*}
which leads to the adjoint equation:

\begin{linenomath*}
\begin{equation}
\left(\frac{\partial \mathbf{r}}{\partial \mathbf{x}}\right)^T \boldsymbol{\lambda} = \left(\frac{\partial J}{\partial \mathbf{x}}\right)^T \label{eq:adjoint_condition_transpose}
\end{equation}
\end{linenomath*}
In the special case of a linear forward model $\mathbf{r}(\mathbf{x},\mathbf{g}) = \mathbf{A}(\mathbf{g})\mathbf{x} - \mathbf{b}$, the partial derivative simplifies to 
\begin{linenomath*}
\begin{equation}
\frac{\partial \mathbf{r}}{\partial \mathbf{x}} = \frac{\partial \mathbf{A}(\mathbf{g}) \mathbf{x}}{\partial \mathbf{x}} = \mathbf{A} \label{eq:drdx_is_A}
\end{equation}
\end{linenomath*}
The adjoint equation becomes:

\begin{linenomath*}
\begin{equation}
\mathbf{A}^T \boldsymbol{\lambda} = \left(\frac{\partial J}{\partial \mathbf{x}}\right)^T \label{eq:adjoint_final}
\end{equation}
\end{linenomath*}
This adjoint equation ensures that the derivative of $\mathbf{x}$ with respect to hydraulic conductivity $g_i$ no longer explicitly appears in the gradient. The final expression for the gradient is then simplified to:

\begin{linenomath*}
\begin{equation}
\frac{dJ}{dg_i} = \frac{\partial J}{\partial g_i} - \boldsymbol{\lambda}^T \frac{\partial \mathbf{r}}{\partial g_i} \label{eq:adjoint_gradient}
\end{equation}
\end{linenomath*}
Since permeability $K$ is a function of the state vector $\mathbf{x}$, i.e., $K = K(\mathbf{x})$, it follows that 

\begin{linenomath*}
\begin{equation}
\frac{\partial J}{\partial g_i} = \frac{\partial J}{\partial K} \frac{\partial K}{\partial g_i} \label{eq:partialJ_partialgi}
\end{equation}
\end{linenomath*}
Substituting Equation~\ref{eq:residual} to Equation~\ref{eq:adjoint_gradient}, this derivative simplifies to
\begin{linenomath*}
\begin{equation}
\frac{dJ}{dg_i} = \frac{\partial J}{\partial g_i} - \boldsymbol{\lambda}^T \frac{\partial \mathbf{A}(g)}{\partial g_i}\mathbf{x} \label{eq:dr_dgi}
\end{equation}
\end{linenomath*}
Finally, by substituting Equation~\ref{eq:partialJ_partialgi} to Equation~\ref{eq:dr_dgi}, we obtain:

\begin{linenomath*}
\begin{equation}
\frac{dJ}{dg_i} = \frac{\partial J}{\partial K} \frac{\partial K}{\partial g_i}  
- \boldsymbol{\lambda}^T \left(\frac{\partial \mathbf{A}}{\partial g_i}\right) \mathbf{x} \label{eq:final_gradient}
\end{equation}
\end{linenomath*}
with the adjoint equation given by Equation~\ref{eq:adjoint_final}. Thus, solving the adjoint equation yields the gradient of loss with respect to hydraulic conductivity $\frac{dJ}{dg_i}$ efficiently without explicitly computing $\frac{d\mathbf{x}}{dg_i}$, making the discrete adjoint approach computationally efficient for parameter estimation and data assimilation in linear systems.

In the adjoint equation Equation~\ref{eq:adjoint_final}, $\mathbf{A}(\mathbf{g}(\mathbf{w}))^{\mathrm{T}}$ is the transpose of the system matrix from the forward problem, and $\frac{\partial J}{\partial \mathbf{x}}$ must be computed from the chain rule using Equations~\ref{eq:KcQ1} and \ref{eq:objfunc}. Equations~\ref{eq:adjoint_final} and~\ref{eq:final_gradient} provide the core formulation for computing the gradient \( dJ/dg_i \) using the adjoint method. To solve Equations~\ref{eq:adjoint_final} and~\ref{eq:final_gradient}, we first determine $\frac{\partial J}{\partial K}$ and $\frac{\partial K}{\partial \mathbf{x}}$:
\begin{linenomath*}
\begin{equation}
\frac{\partial J}{\partial K} = K - K^*
\label{eq:dJdK}
\end{equation}
\end{linenomath*}

\begin{linenomath*}
\begin{equation}
\frac{\partial K}{\partial \mathbf{x}} = c\,\frac{\partial Q_{\mathrm{in}}}{\partial \mathbf{x}} 
\label{eq:dKdx}
\end{equation}
\end{linenomath*}
Multiplying these gives
\begin{linenomath*}
\begin{equation}
\frac{\partial J}{\partial \mathbf{x}} = \frac{\partial J}{\partial K}\frac{\partial K}{\partial \mathbf{x}} = (K - K^*) c\,\frac{\partial Q_{\mathrm{in}}}{\partial \mathbf{x}}  \label{eq:dJdx}
\end{equation}
\end{linenomath*}

The derivative for the inlet flow rate $Q_{\mathrm{in}}$ involves only the inlet pores and the throats connected to them, we define vector $\mathbf{L_Q}$ such that:
\begin{linenomath*}
\begin{equation}
Q_{\mathrm{in}} = \mathbf{L_Q}^T\mathbf{x}. \label{eq:L_Q}
\end{equation}
\end{linenomath*}
The vector $\mathbf{L_Q}$ has non-zero entries only at pores connected to inlet throats. Substitute Equation~\ref{eq:dJdx} into Equation~\ref{eq:adjoint_final} to solve for Lagrange multiplier $\boldsymbol{\lambda}$. Once Lagrange multiplier $\boldsymbol{\lambda}$ is known, we return to Equation~\ref{eq:final_gradient} to evaluate $\frac{\partial J}{\partial g_i}$ for each throat. The rest part of term $\boldsymbol{\lambda}^{\mathrm{T}} \left(\frac{\partial \mathbf{A}}{\partial g_i}\right) \mathbf{x}$ in Equation~\ref{eq:final_gradient} can be solved easily. For a throat $i$ that connects two pores $p_1$ and $p_2$, the matrix $\mathbf{A}$ has nonzero entries only at the $(p_1,p_1)$, $(p_2,p_2)$, $(p_1,p_2)$, and $(p_2,p_1)$ locations. Consequently,
\begin{linenomath*}
\begin{equation}
\frac{\partial A_{p_1,p_1}}{\partial g_i} = 1,\quad
\frac{\partial A_{p_2,p_2}}{\partial g_i} = 1,\quad
\frac{\partial A_{p_1,p_2}}{\partial g_i} = -1,\quad
\frac{\partial A_{p_2,p_1}}{\partial g_i} = -1. \label{eq:dAdg}
\end{equation}
\end{linenomath*}
and all other entries are zero. Multiplying this sparse pattern by the state vector $\mathbf{x}$ and dotting with $-\boldsymbol{\lambda}^{\mathrm{T}}$ yields
\begin{linenomath*}
\begin{equation}
-\boldsymbol{\lambda}^{\mathrm{T}} \left(\frac{\partial \mathbf{A}}{\partial g_i}\right) \mathbf{x} 
= (\lambda_{p_2} - \lambda_{p_1})(x_{p_1} - x_{p_2}). \label{eq:adjoint_gradient}
\end{equation}
\end{linenomath*}

For the second term in Equation~\ref{eq:dJdgi-general},  the first part $\frac{\partial J}{\partial K}$ has already been calculated by Equation~\ref{eq:dJdK}. Meanwhile,
\begin{linenomath*}
\begin{equation}
\frac{\partial K}{\partial g_i} = c\,\frac{\partial Q_{\mathrm{in}}}{\partial g_i}. \label{eq:dK_dgi}
\end{equation}
\end{linenomath*}

Because the inlet flow rate $Q_{\mathrm{in}}$ is the net flow entering the inlet pores, only those throats that connect directly to an inlet pore can change the inlet flow rate $Q_{\mathrm{in}}$. In that case,
\begin{linenomath*}
\begin{equation}
\frac{\partial Q_{\mathrm{in}}}{\partial g_i} = x_{p_1} - x_{p_2}. \label{eq:dQin_dgi}
\end{equation}
\end{linenomath*}
otherwise it is zero. Hence,
\begin{linenomath*}
\begin{equation}
\frac{\partial J}{\partial K}\frac{\partial K}{\partial g_i} 
= (K - K^*)\, c\, \delta_{\mathrm{inlet}}(i)(x_{p_1} - x_{p_2}). \label{eq:dJdK_dKdg}
\end{equation}
\end{linenomath*}
where $\delta_{\mathrm{inlet}}(i) = 1$ if throat $i$ meets an inlet pore and $0$ otherwise. Putting both terms together, we obtain
\begin{linenomath*}
\begin{equation}
\frac{d J}{d g_i} = (\lambda_{p_2} - \lambda_{p_1})(x_{p_1} - x_{p_2}) + (K - K^*) c \delta_{\mathrm{inlet}}(i)(x_{p_1} - x_{p_2}).
\label{eq:dJdgi-final-app}
\end{equation}
\end{linenomath*}

An alternative method for computing the gradient of the loss \(J\) with respect to conductance \(g_i\) is the finite difference (FD) approach. In finite difference, each conductance \(g_i\) is perturbed individually by a small amount \(\Delta g_i\), and the resulting change in the loss function \(J\) is used to approximate the gradient:
\begin{linenomath*}
\begin{equation}
\biggl(\tfrac{d J}{d g_i}\biggr)^{\mathrm{FD}}
\;=\;
\frac{J\!\bigl(\mathbf{g}^{\,b} + \Delta g_i\, \mathbf{e}_i\bigr)\;-\;J\!\bigl(\mathbf{g}^{\,b}\bigr)}
     {\Delta g_i},
\label{eq:fd-approx}
\end{equation}
\end{linenomath*}
where \( \mathbf{g}^{\,b} = (g_{1},\,\dots,\,g_{i},\,\dots,\,g_{N_e}) \) is the baseline conductance vector, and \( \mathbf{e}_i \) is the unit vector in the \(i\)-th direction. The perturbed vector differs from baseline conductance vector \(\mathbf{g}^{\,b}\) only in the \(i\)-th component.

While straightforward to implement, the finite-difference approach requires a separate forward model evaluation for each parameter \(g_i\). This becomes computationally prohibitive in high-dimensional settings, such as pore network modeling with thousands of conductance quantities.
In contrast, the discrete adjoint method computes all gradients of the loss with respect to the hydraulic conductance \(\tfrac{d J}{d g_i}\) simultaneously by solving a single linear system whose size depends only on the number of pores—not on the number of conductances. This is particularly efficient in our framework, where a large set of conductances determines a single scalar output (permeability). The adjoint approach is thus highly scalable and mirrors efficient strategies commonly used in PDE-constrained optimization, where one seeks to infer high-dimensional parameters from limited scalar observations.

\bibliography{ref}

\begin{thebibliography}{}

\bibitem [\protect \citeauthoryear {%
Adeyemi%
, Ghanbarian%
, Winter%
\BCBL {}\ \BBA {} King%
}{%
Adeyemi%
\ \protect \BOthers {.}}{%
{\protect \APACyear {2022}}%
}]{%
adeyemi2022determining}
\APACinsertmetastar {%
adeyemi2022determining}%
\begin{APACrefauthors}%
Adeyemi, B.%
, Ghanbarian, B.%
, Winter, C.%
\BCBL {}\ \BBA {} King, P\BPBI R.%
\end{APACrefauthors}%
\unskip\
\newblock
\APACrefYearMonthDay{2022}{}{}.
\newblock
{\BBOQ}\APACrefatitle {Determining effective permeability at reservoir scale:
  Application of critical path analysis} {Determining effective permeability at
  reservoir scale: Application of critical path analysis}.{\BBCQ}
\newblock
\APACjournalVolNumPages{Advances in Water Resources}{159}{}{104096}.
\PrintBackRefs{\CurrentBib}

\bibitem [\protect \citeauthoryear {%
Alzahrani%
, Shapoval%
, Chen%
\BCBL {}\ \BBA {} Rahman%
}{%
Alzahrani%
\ \protect \BOthers {.}}{%
{\protect \APACyear {2023}}%
}]{%
alzahrani2023pore}
\APACinsertmetastar {%
alzahrani2023pore}%
\begin{APACrefauthors}%
Alzahrani, M\BPBI K.%
, Shapoval, A.%
, Chen, Z.%
\BCBL {}\ \BBA {} Rahman, S\BPBI S.%
\end{APACrefauthors}%
\unskip\
\newblock
\APACrefYearMonthDay{2023}{}{}.
\newblock
{\BBOQ}\APACrefatitle {Pore-GNN: A graph neural network-based framework for
  predicting flow properties of porous media from micro-CT images} {Pore-gnn: A
  graph neural network-based framework for predicting flow properties of porous
  media from micro-ct images}.{\BBCQ}
\newblock
\APACjournalVolNumPages{Advances in Geo-Energy Research}{10}{1}{39--55}.
\PrintBackRefs{\CurrentBib}

\bibitem [\protect \citeauthoryear {%
Bickle%
}{%
Bickle%
}{%
{\protect \APACyear {2009}}%
}]{%
bickle2009geological}
\APACinsertmetastar {%
bickle2009geological}%
\begin{APACrefauthors}%
Bickle, M\BPBI J.%
\end{APACrefauthors}%
\unskip\
\newblock
\APACrefYearMonthDay{2009}{}{}.
\newblock
{\BBOQ}\APACrefatitle {Geological carbon storage} {Geological carbon
  storage}.{\BBCQ}
\newblock
\APACjournalVolNumPages{Nature Geoscience}{2}{12}{815--818}.
\PrintBackRefs{\CurrentBib}

\bibitem [\protect \citeauthoryear {%
Blunt%
}{%
Blunt%
}{%
{\protect \APACyear {2001}}%
}]{%
blunt2001flow}
\APACinsertmetastar {%
blunt2001flow}%
\begin{APACrefauthors}%
Blunt, M\BPBI J.%
\end{APACrefauthors}%
\unskip\
\newblock
\APACrefYearMonthDay{2001}{}{}.
\newblock
{\BBOQ}\APACrefatitle {Flow in porous media—pore-network models and
  multiphase flow} {Flow in porous media—pore-network models and multiphase
  flow}.{\BBCQ}
\newblock
\APACjournalVolNumPages{Current Opinion in Colloid \& Interface
  Science}{6}{3}{197--207}.
\PrintBackRefs{\CurrentBib}

\bibitem [\protect \citeauthoryear {%
Brenner%
}{%
Brenner%
}{%
{\protect \APACyear {2024}}%
}]{%
brenner2024variational}
\APACinsertmetastar {%
brenner2024variational}%
\begin{APACrefauthors}%
Brenner, O.%
\end{APACrefauthors}%
\unskip\
\newblock
\APACrefYear{2024}.
\unskip\
\newblock
\APACrefbtitle {A Variational Data Assimilation Approach for Stationary
  Turbulent Flow Simulations} {A variational data assimilation approach for
  stationary turbulent flow simulations}\ \APACtypeAddressSchool {\BUPhD}{}{}.
\unskip\
\newblock
\APACaddressSchool {}{ETH Zurich}.
\PrintBackRefs{\CurrentBib}

\bibitem [\protect \citeauthoryear {%
Caglar%
, Broggi%
, Ali%
, Org{\'e}as%
\BCBL {}\ \BBA {} Michaud%
}{%
Caglar%
\ \protect \BOthers {.}}{%
{\protect \APACyear {2022}}%
}]{%
caglar2022deep}
\APACinsertmetastar {%
caglar2022deep}%
\begin{APACrefauthors}%
Caglar, B.%
, Broggi, G.%
, Ali, M\BPBI A.%
, Org{\'e}as, L.%
\BCBL {}\ \BBA {} Michaud, V.%
\end{APACrefauthors}%
\unskip\
\newblock
\APACrefYearMonthDay{2022}{}{}.
\newblock
{\BBOQ}\APACrefatitle {Deep learning accelerated prediction of the permeability
  of fibrous microstructures} {Deep learning accelerated prediction of the
  permeability of fibrous microstructures}.{\BBCQ}
\newblock
\APACjournalVolNumPages{Composites Part A: Applied Science and
  Manufacturing}{158}{}{106973}.
\PrintBackRefs{\CurrentBib}

\bibitem [\protect \citeauthoryear {%
Corso%
, Stark%
, Jegelka%
, Jaakkola%
\BCBL {}\ \BBA {} Barzilay%
}{%
Corso%
\ \protect \BOthers {.}}{%
{\protect \APACyear {2024}}%
}]{%
corso2024graph}
\APACinsertmetastar {%
corso2024graph}%
\begin{APACrefauthors}%
Corso, G.%
, Stark, H.%
, Jegelka, S.%
, Jaakkola, T.%
\BCBL {}\ \BBA {} Barzilay, R.%
\end{APACrefauthors}%
\unskip\
\newblock
\APACrefYearMonthDay{2024}{}{}.
\newblock
{\BBOQ}\APACrefatitle {Graph neural networks} {Graph neural networks}.{\BBCQ}
\newblock
\APACjournalVolNumPages{Nature Reviews Methods Primers}{4}{1}{17}.
\PrintBackRefs{\CurrentBib}

\bibitem [\protect \citeauthoryear {%
dos Anjos%
\ \protect \BOthers {.}}{%
dos Anjos%
\ \protect \BOthers {.}}{%
{\protect \APACyear {2023}}%
}]{%
dos2023permeability}
\APACinsertmetastar {%
dos2023permeability}%
\begin{APACrefauthors}%
dos Anjos, C\BPBI E\BPBI M.%
, de Matos, T\BPBI F.%
, Avila, M\BPBI R\BPBI V.%
, Fernandes, J\BPBI d\BPBI C\BPBI V.%
, Surmas, R.%
\BCBL {}\ \BBA {} Evsukoff, A\BPBI G.%
\end{APACrefauthors}%
\unskip\
\newblock
\APACrefYearMonthDay{2023}{}{}.
\newblock
{\BBOQ}\APACrefatitle {Permeability estimation on raw micro-{CT} of carbonate
  rock samples using deep learning} {Permeability estimation on raw micro-{CT}
  of carbonate rock samples using deep learning}.{\BBCQ}
\newblock
\APACjournalVolNumPages{Geoenergy Science and Engineering}{222}{}{211335}.
\PrintBackRefs{\CurrentBib}

\bibitem [\protect \citeauthoryear {%
Duraisamy%
, Iaccarino%
\BCBL {}\ \BBA {} Xiao%
}{%
Duraisamy%
\ \protect \BOthers {.}}{%
{\protect \APACyear {2019}}%
}]{%
duraisamy2019turbulence}
\APACinsertmetastar {%
duraisamy2019turbulence}%
\begin{APACrefauthors}%
Duraisamy, K.%
, Iaccarino, G.%
\BCBL {}\ \BBA {} Xiao, H.%
\end{APACrefauthors}%
\unskip\
\newblock
\APACrefYearMonthDay{2019}{}{}.
\newblock
{\BBOQ}\APACrefatitle {Turbulence modeling in the age of data} {Turbulence
  modeling in the age of data}.{\BBCQ}
\newblock
\APACjournalVolNumPages{Annual Review of Fluid Mechanics}{51}{1}{357--377}.
\PrintBackRefs{\CurrentBib}

\bibitem [\protect \citeauthoryear {%
Elmorsy%
, El-Dakhakhni%
\BCBL {}\ \BBA {} Zhao%
}{%
Elmorsy%
\ \protect \BOthers {.}}{%
{\protect \APACyear {2022}}%
}]{%
elmorsy2022generalizable}
\APACinsertmetastar {%
elmorsy2022generalizable}%
\begin{APACrefauthors}%
Elmorsy, M.%
, El-Dakhakhni, W.%
\BCBL {}\ \BBA {} Zhao, B.%
\end{APACrefauthors}%
\unskip\
\newblock
\APACrefYearMonthDay{2022}{}{}.
\newblock
{\BBOQ}\APACrefatitle {Generalizable permeability prediction of digital porous
  media via a novel multi-scale 3D convolutional neural network} {Generalizable
  permeability prediction of digital porous media via a novel multi-scale 3d
  convolutional neural network}.{\BBCQ}
\newblock
\APACjournalVolNumPages{Water Resources Research}{58}{3}{e2021WR031454}.
\PrintBackRefs{\CurrentBib}

\bibitem [\protect \citeauthoryear {%
Elmorsy%
, El-Dakhakhni%
\BCBL {}\ \BBA {} Zhao%
}{%
Elmorsy%
\ \protect \BOthers {.}}{%
{\protect \APACyear {2023}}%
}]{%
elmorsy2023rapid}
\APACinsertmetastar {%
elmorsy2023rapid}%
\begin{APACrefauthors}%
Elmorsy, M.%
, El-Dakhakhni, W.%
\BCBL {}\ \BBA {} Zhao, B.%
\end{APACrefauthors}%
\unskip\
\newblock
\APACrefYearMonthDay{2023}{}{}.
\newblock
{\BBOQ}\APACrefatitle {Rapid permeability upscaling of digital porous media via
  physics-informed neural networks} {Rapid permeability upscaling of digital
  porous media via physics-informed neural networks}.{\BBCQ}
\newblock
\APACjournalVolNumPages{Water Resources Research}{59}{12}{e2023WR035064}.
\PrintBackRefs{\CurrentBib}

\bibitem [\protect \citeauthoryear {%
Erickson%
, Korfiatis%
, Akkus%
\BCBL {}\ \BBA {} Kline%
}{%
Erickson%
\ \protect \BOthers {.}}{%
{\protect \APACyear {2017}}%
}]{%
erickson2017machine}
\APACinsertmetastar {%
erickson2017machine}%
\begin{APACrefauthors}%
Erickson, B\BPBI J.%
, Korfiatis, P.%
, Akkus, Z.%
\BCBL {}\ \BBA {} Kline, T\BPBI L.%
\end{APACrefauthors}%
\unskip\
\newblock
\APACrefYearMonthDay{2017}{}{}.
\newblock
{\BBOQ}\APACrefatitle {Machine learning for medical imaging} {Machine learning
  for medical imaging}.{\BBCQ}
\newblock
\APACjournalVolNumPages{Radiographics}{37}{2}{505--515}.
\PrintBackRefs{\CurrentBib}

\bibitem [\protect \citeauthoryear {%
Fu%
\ \protect \BOthers {.}}{%
Fu%
\ \protect \BOthers {.}}{%
{\protect \APACyear {2023}}%
}]{%
fu2023data}
\APACinsertmetastar {%
fu2023data}%
\begin{APACrefauthors}%
Fu, J.%
, Wang, M.%
, Chen, B.%
, Wang, J.%
, Xiao, D.%
, Luo, M.%
\BCBL {}\ \BBA {} Evans, B.%
\end{APACrefauthors}%
\unskip\
\newblock
\APACrefYearMonthDay{2023}{}{}.
\newblock
{\BBOQ}\APACrefatitle {A data-driven framework for permeability prediction of
  natural porous rocks via microstructural characterization and pore-scale
  simulation} {A data-driven framework for permeability prediction of natural
  porous rocks via microstructural characterization and pore-scale
  simulation}.{\BBCQ}
\newblock
\APACjournalVolNumPages{Engineering with Computers}{39}{6}{3895--3926}.
\PrintBackRefs{\CurrentBib}

\bibitem [\protect \citeauthoryear {%
G{\"a}rttner%
, Alpak%
, Meier%
, Ray%
\BCBL {}\ \BBA {} Frank%
}{%
G{\"a}rttner%
\ \protect \BOthers {.}}{%
{\protect \APACyear {2023}}%
}]{%
garttner2023estimating}
\APACinsertmetastar {%
garttner2023estimating}%
\begin{APACrefauthors}%
G{\"a}rttner, S.%
, Alpak, F\BPBI O.%
, Meier, A.%
, Ray, N.%
\BCBL {}\ \BBA {} Frank, F.%
\end{APACrefauthors}%
\unskip\
\newblock
\APACrefYearMonthDay{2023}{}{}.
\newblock
{\BBOQ}\APACrefatitle {Estimating permeability of {3D} micro-{CT} images by
  physics-informed {CNNs} based on {DNS}} {Estimating permeability of {3D}
  micro-{CT} images by physics-informed {CNNs} based on {DNS}}.{\BBCQ}
\newblock
\APACjournalVolNumPages{Computational Geosciences}{27}{2}{245--262}.
\PrintBackRefs{\CurrentBib}

\bibitem [\protect \citeauthoryear {%
J.~Gostick%
\ \protect \BOthers {.}}{%
J.~Gostick%
\ \protect \BOthers {.}}{%
{\protect \APACyear {2016}}%
}]{%
gostick2016openpnm}
\APACinsertmetastar {%
gostick2016openpnm}%
\begin{APACrefauthors}%
Gostick, J.%
, Aghighi, M.%
, Hinebaugh, J.%
, Tranter, T.%
, Hoeh, M\BPBI A.%
, Day, H.%
\BDBL {}others%
\end{APACrefauthors}%
\unskip\
\newblock
\APACrefYearMonthDay{2016}{}{}.
\newblock
{\BBOQ}\APACrefatitle {OpenPNM: a pore network modeling package} {Openpnm: a
  pore network modeling package}.{\BBCQ}
\newblock
\APACjournalVolNumPages{Computing in Science \& Engineering}{18}{4}{60--74}.
\PrintBackRefs{\CurrentBib}

\bibitem [\protect \citeauthoryear {%
J\BPBI T.~Gostick%
\ \protect \BOthers {.}}{%
J\BPBI T.~Gostick%
\ \protect \BOthers {.}}{%
{\protect \APACyear {2019}}%
}]{%
gostick2019porespy}
\APACinsertmetastar {%
gostick2019porespy}%
\begin{APACrefauthors}%
Gostick, J\BPBI T.%
, Khan, Z\BPBI A.%
, Tranter, T\BPBI G.%
, Kok, M\BPBI D.%
, Agnaou, M.%
, Sadeghi, M.%
\BCBL {}\ \BBA {} Jervis, R.%
\end{APACrefauthors}%
\unskip\
\newblock
\APACrefYearMonthDay{2019}{}{}.
\newblock
{\BBOQ}\APACrefatitle {PoreSpy: A python toolkit for quantitative analysis of
  porous media images} {Porespy: A python toolkit for quantitative analysis of
  porous media images}.{\BBCQ}
\newblock
\APACjournalVolNumPages{Journal of Open Source Software}{4}{37}{1296}.
\PrintBackRefs{\CurrentBib}

\bibitem [\protect \citeauthoryear {%
Guo%
\ \protect \BOthers {.}}{%
Guo%
\ \protect \BOthers {.}}{%
{\protect \APACyear {2022}}%
}]{%
guo2022role}
\APACinsertmetastar {%
guo2022role}%
\begin{APACrefauthors}%
Guo, R.%
, Dalton, L.%
, Crandall, D.%
, McClure, J.%
, Wang, H.%
, Li, Z.%
\BCBL {}\ \BBA {} Chen, C.%
\end{APACrefauthors}%
\unskip\
\newblock
\APACrefYearMonthDay{2022}{}{}.
\newblock
{\BBOQ}\APACrefatitle {Role of heterogeneous surface wettability on dynamic
  immiscible displacement, capillary pressure, and relative permeability in a
  CO2-water-rock system} {Role of heterogeneous surface wettability on dynamic
  immiscible displacement, capillary pressure, and relative permeability in a
  co2-water-rock system}.{\BBCQ}
\newblock
\APACjournalVolNumPages{Advances in Water Resources}{165}{}{104226}.
\PrintBackRefs{\CurrentBib}

\bibitem [\protect \citeauthoryear {%
Han%
, Feng%
\BCBL {}\ \BBA {} Ning%
}{%
Han%
\ \protect \BOthers {.}}{%
{\protect \APACyear {2024}}%
}]{%
han2024topology}
\APACinsertmetastar {%
han2024topology}%
\begin{APACrefauthors}%
Han, X.%
, Feng, Z.%
\BCBL {}\ \BBA {} Ning, Y.%
\end{APACrefauthors}%
\unskip\
\newblock
\APACrefYearMonthDay{2024}{}{}.
\newblock
{\BBOQ}\APACrefatitle {A topology-aware graph coarsening framework for
  continual graph learning} {A topology-aware graph coarsening framework for
  continual graph learning}.{\BBCQ}
\newblock
\APACjournalVolNumPages{Advances in Neural Information Processing
  Systems}{37}{}{132491--132523}.
\PrintBackRefs{\CurrentBib}

\bibitem [\protect \citeauthoryear {%
Jameson%
}{%
Jameson%
}{%
{\protect \APACyear {1988}}%
}]{%
jameson1988aerodynamic}
\APACinsertmetastar {%
jameson1988aerodynamic}%
\begin{APACrefauthors}%
Jameson, A.%
\end{APACrefauthors}%
\unskip\
\newblock
\APACrefYearMonthDay{1988}{}{}.
\newblock
{\BBOQ}\APACrefatitle {Aerodynamic design via control theory} {Aerodynamic
  design via control theory}.{\BBCQ}
\newblock
\APACjournalVolNumPages{Journal of scientific computing}{3}{}{233--260}.
\PrintBackRefs{\CurrentBib}

\bibitem [\protect \citeauthoryear {%
Jiang%
\ \protect \BOthers {.}}{%
Jiang%
\ \protect \BOthers {.}}{%
{\protect \APACyear {2023}}%
}]{%
jiang2023upscaling}
\APACinsertmetastar {%
jiang2023upscaling}%
\begin{APACrefauthors}%
Jiang, F.%
, Guo, Y.%
, Tsuji, T.%
, Kato, Y.%
, Shimokawara, M.%
, Esteban, L.%
\BDBL {}Kitamura, R.%
\end{APACrefauthors}%
\unskip\
\newblock
\APACrefYearMonthDay{2023}{}{}.
\newblock
{\BBOQ}\APACrefatitle {Upscaling permeability using multiscale {X}-ray-{CT}
  images with digital rock modeling and deep learning techniques} {Upscaling
  permeability using multiscale {X}-ray-{CT} images with digital rock modeling
  and deep learning techniques}.{\BBCQ}
\newblock
\APACjournalVolNumPages{Water Resources Research}{59}{3}{e2022WR033267}.
\PrintBackRefs{\CurrentBib}

\bibitem [\protect \citeauthoryear {%
Joekar-Niasar%
\ \BBA {} Hassanizadeh%
}{%
Joekar-Niasar%
\ \BBA {} Hassanizadeh%
}{%
{\protect \APACyear {2012}}%
}]{%
joekar2012analysis}
\APACinsertmetastar {%
joekar2012analysis}%
\begin{APACrefauthors}%
Joekar-Niasar, V.%
\BCBT {}\ \BBA {} Hassanizadeh, S.%
\end{APACrefauthors}%
\unskip\
\newblock
\APACrefYearMonthDay{2012}{}{}.
\newblock
{\BBOQ}\APACrefatitle {Analysis of fundamentals of two-phase flow in porous
  media using dynamic pore-network models: A review} {Analysis of fundamentals
  of two-phase flow in porous media using dynamic pore-network models: A
  review}.{\BBCQ}
\newblock
\APACjournalVolNumPages{Critical reviews in environmental science and
  technology}{42}{18}{1895--1976}.
\PrintBackRefs{\CurrentBib}

\bibitem [\protect \citeauthoryear {%
Jones%
, Smith%
\BCBL {}\ \BBA {} Lee%
}{%
Jones%
\ \protect \BOthers {.}}{%
{\protect \APACyear {2021}}%
}]{%
jones2021improved}
\APACinsertmetastar {%
jones2021improved}%
\begin{APACrefauthors}%
Jones, D.%
, Smith, A.%
\BCBL {}\ \BBA {} Lee, M.%
\end{APACrefauthors}%
\unskip\
\newblock
\APACrefYearMonthDay{2021}{}{}.
\newblock
{\BBOQ}\APACrefatitle {Improved protein--ligand binding affinity prediction
  with structure-based deep fusion inference} {Improved protein--ligand binding
  affinity prediction with structure-based deep fusion inference}.{\BBCQ}
\newblock
\APACjournalVolNumPages{Journal of Chemical Information and
  Modeling}{61}{4}{1583--1592}.
\PrintBackRefs{\CurrentBib}

\bibitem [\protect \citeauthoryear {%
Kamrava%
, Tahmasebi%
\BCBL {}\ \BBA {} Sahimi%
}{%
Kamrava%
\ \protect \BOthers {.}}{%
{\protect \APACyear {2020}}%
}]{%
kamrava2020linking}
\APACinsertmetastar {%
kamrava2020linking}%
\begin{APACrefauthors}%
Kamrava, S.%
, Tahmasebi, P.%
\BCBL {}\ \BBA {} Sahimi, M.%
\end{APACrefauthors}%
\unskip\
\newblock
\APACrefYearMonthDay{2020}{}{}.
\newblock
{\BBOQ}\APACrefatitle {Linking morphology of porous media to their macroscopic
  permeability by deep learning} {Linking morphology of porous media to their
  macroscopic permeability by deep learning}.{\BBCQ}
\newblock
\APACjournalVolNumPages{Transport in Porous Media}{131}{2}{427--448}.
\PrintBackRefs{\CurrentBib}

\bibitem [\protect \citeauthoryear {%
Kang%
, Li%
, Fu%
\BCBL {}\ \BBA {} Liu%
}{%
Kang%
\ \protect \BOthers {.}}{%
{\protect \APACyear {2024}}%
}]{%
kang2024hybrid}
\APACinsertmetastar {%
kang2024hybrid}%
\begin{APACrefauthors}%
Kang, Q.%
, Li, K\BHBI Q.%
, Fu, J\BHBI L.%
\BCBL {}\ \BBA {} Liu, Y.%
\end{APACrefauthors}%
\unskip\
\newblock
\APACrefYearMonthDay{2024}{}{}.
\newblock
{\BBOQ}\APACrefatitle {Hybrid {LBM} and machine learning algorithms for
  permeability prediction of porous media: a comparative study} {Hybrid {LBM}
  and machine learning algorithms for permeability prediction of porous media:
  a comparative study}.{\BBCQ}
\newblock
\APACjournalVolNumPages{Computers and Geotechnics}{168}{}{106163}.
\PrintBackRefs{\CurrentBib}

\bibitem [\protect \citeauthoryear {%
Kashefi%
\ \BBA {} Mukerji%
}{%
Kashefi%
\ \BBA {} Mukerji%
}{%
{\protect \APACyear {2021}}%
}]{%
kashefi2021point}
\APACinsertmetastar {%
kashefi2021point}%
\begin{APACrefauthors}%
Kashefi, A.%
\BCBT {}\ \BBA {} Mukerji, T.%
\end{APACrefauthors}%
\unskip\
\newblock
\APACrefYearMonthDay{2021}{}{}.
\newblock
{\BBOQ}\APACrefatitle {Point-cloud deep learning of porous media for
  permeability prediction} {Point-cloud deep learning of porous media for
  permeability prediction}.{\BBCQ}
\newblock
\APACjournalVolNumPages{Physics of Fluids}{33}{9}{}.
\PrintBackRefs{\CurrentBib}

\bibitem [\protect \citeauthoryear {%
Kim%
\ \BBA {} Suh%
}{%
Kim%
\ \BBA {} Suh%
}{%
{\protect \APACyear {2025}}%
}]{%
KIM2025107497}
\APACinsertmetastar {%
KIM2025107497}%
\begin{APACrefauthors}%
Kim, Y.%
\BCBT {}\ \BBA {} Suh, H\BPBI S.%
\end{APACrefauthors}%
\unskip\
\newblock
\APACrefYearMonthDay{2025}{}{}.
\newblock
{\BBOQ}\APACrefatitle {{GNPNM}: A graph neural pore network model for
  predicting quasi-static drainage displacement patterns} {{GNPNM}: A graph
  neural pore network model for predicting quasi-static drainage displacement
  patterns}.{\BBCQ}
\newblock
\APACjournalVolNumPages{Computers and Geotechnics}{187}{}{107497}.
\newblock
\begin{APACrefURL}
  \url{https://www.sciencedirect.com/science/article/pii/S0266352X2500446X}
  \end{APACrefURL}
\newblock
\begin{APACrefDOI} \doi{https://doi.org/10.1016/j.compgeo.2025.107497}
  \end{APACrefDOI}
\PrintBackRefs{\CurrentBib}

\bibitem [\protect \citeauthoryear {%
S.~Liu%
, Fan%
\BCBL {}\ \BBA {} Lu%
}{%
S.~Liu%
\ \protect \BOthers {.}}{%
{\protect \APACyear {2023}}%
}]{%
liu2023uncertainty}
\APACinsertmetastar {%
liu2023uncertainty}%
\begin{APACrefauthors}%
Liu, S.%
, Fan, M.%
\BCBL {}\ \BBA {} Lu, D.%
\end{APACrefauthors}%
\unskip\
\newblock
\APACrefYearMonthDay{2023}{}{}.
\newblock
{\BBOQ}\APACrefatitle {Uncertainty quantification of the convolutional neural
  networks on permeability estimation from micro-CT scanned sandstone and
  carbonate rock images} {Uncertainty quantification of the convolutional
  neural networks on permeability estimation from micro-ct scanned sandstone
  and carbonate rock images}.{\BBCQ}
\newblock
\APACjournalVolNumPages{Geoenergy Science and Engineering}{230}{}{212160}.
\PrintBackRefs{\CurrentBib}

\bibitem [\protect \citeauthoryear {%
Y.~Liu%
, Gong%
, Zhao%
, Jin%
\BCBL {}\ \BBA {} Wang%
}{%
Y.~Liu%
\ \protect \BOthers {.}}{%
{\protect \APACyear {2022}}%
}]{%
liu2022pore}
\APACinsertmetastar {%
liu2022pore}%
\begin{APACrefauthors}%
Liu, Y.%
, Gong, W.%
, Zhao, Y.%
, Jin, X.%
\BCBL {}\ \BBA {} Wang, M.%
\end{APACrefauthors}%
\unskip\
\newblock
\APACrefYearMonthDay{2022}{}{}.
\newblock
{\BBOQ}\APACrefatitle {A pore-throat segmentation method based on local
  hydraulic resistance equivalence for pore-network modeling} {A pore-throat
  segmentation method based on local hydraulic resistance equivalence for
  pore-network modeling}.{\BBCQ}
\newblock
\APACjournalVolNumPages{Water Resources Research}{58}{12}{e2022WR033142}.
\PrintBackRefs{\CurrentBib}

\bibitem [\protect \citeauthoryear {%
Martins%
\ \BBA {} Ning%
}{%
Martins%
\ \BBA {} Ning%
}{%
{\protect \APACyear {2021}}%
}]{%
martins2021engineering}
\APACinsertmetastar {%
martins2021engineering}%
\begin{APACrefauthors}%
Martins, J\BPBI R.%
\BCBT {}\ \BBA {} Ning, A.%
\end{APACrefauthors}%
\unskip\
\newblock
\APACrefYear{2021}.
\newblock
\APACrefbtitle {Engineering design optimization} {Engineering design
  optimization}.
\newblock
\APACaddressPublisher{}{Cambridge University Press}.
\PrintBackRefs{\CurrentBib}

\bibitem [\protect \citeauthoryear {%
McClure%
, Prins%
\BCBL {}\ \BBA {} Miller%
}{%
McClure%
\ \protect \BOthers {.}}{%
{\protect \APACyear {2014}}%
}]{%
mcclure2014novel}
\APACinsertmetastar {%
mcclure2014novel}%
\begin{APACrefauthors}%
McClure, J\BPBI E.%
, Prins, J\BPBI F.%
\BCBL {}\ \BBA {} Miller, C\BPBI T.%
\end{APACrefauthors}%
\unskip\
\newblock
\APACrefYearMonthDay{2014}{}{}.
\newblock
{\BBOQ}\APACrefatitle {A novel heterogeneous algorithm to simulate multiphase
  flow in porous media on multicore CPU–GPU systems} {A novel heterogeneous
  algorithm to simulate multiphase flow in porous media on multicore cpu–gpu
  systems}.{\BBCQ}
\newblock
\APACjournalVolNumPages{Computer Physics Communications}{185}{7}{1865--1874}.
\newblock
\begin{APACrefDOI} \doi{10.1016/j.cpc.2014.03.012} \end{APACrefDOI}
\PrintBackRefs{\CurrentBib}

\bibitem [\protect \citeauthoryear {%
Meng%
, Jiang%
, Wu%
\BCBL {}\ \BBA {} Wang%
}{%
Meng%
\ \protect \BOthers {.}}{%
{\protect \APACyear {2023}}%
}]{%
meng2023transformer}
\APACinsertmetastar {%
meng2023transformer}%
\begin{APACrefauthors}%
Meng, Y.%
, Jiang, J.%
, Wu, J.%
\BCBL {}\ \BBA {} Wang, D.%
\end{APACrefauthors}%
\unskip\
\newblock
\APACrefYearMonthDay{2023}{}{}.
\newblock
{\BBOQ}\APACrefatitle {Transformer-based deep learning models for predicting
  permeability of porous media} {Transformer-based deep learning models for
  predicting permeability of porous media}.{\BBCQ}
\newblock
\APACjournalVolNumPages{Advances in Water Resources}{179}{}{104520}.
\PrintBackRefs{\CurrentBib}

\bibitem [\protect \citeauthoryear {%
Michel{\'e}n-Str{\"o}fer%
\ \BBA {} Xiao%
}{%
Michel{\'e}n-Str{\"o}fer%
\ \BBA {} Xiao%
}{%
{\protect \APACyear {2021}}%
}]{%
strofer2021end}
\APACinsertmetastar {%
strofer2021end}%
\begin{APACrefauthors}%
Michel{\'e}n-Str{\"o}fer, C\BPBI A.%
\BCBT {}\ \BBA {} Xiao, H.%
\end{APACrefauthors}%
\unskip\
\newblock
\APACrefYearMonthDay{2021}{}{}.
\newblock
{\BBOQ}\APACrefatitle {End-to-end differentiable learning of turbulence models
  from indirect observations} {End-to-end differentiable learning of turbulence
  models from indirect observations}.{\BBCQ}
\newblock
\APACjournalVolNumPages{Theoretical and Applied Mechanics
  Letters}{11}{4}{100280}.
\PrintBackRefs{\CurrentBib}

\bibitem [\protect \citeauthoryear {%
Misaghian%
\ \protect \BOthers {.}}{%
Misaghian%
\ \protect \BOthers {.}}{%
{\protect \APACyear {2022}}%
}]{%
misaghian2022prediction}
\APACinsertmetastar {%
misaghian2022prediction}%
\begin{APACrefauthors}%
Misaghian, N.%
, Agnaou, M.%
, Sadeghi, M\BPBI A.%
, Fathiannasab, H.%
, Hadji, I.%
, Roberts, E.%
\BCBL {}\ \BBA {} Gostick, J.%
\end{APACrefauthors}%
\unskip\
\newblock
\APACrefYearMonthDay{2022}{}{}.
\newblock
{\BBOQ}\APACrefatitle {Prediction of diffusional conductance in extracted pore
  network models using convolutional neural networks} {Prediction of
  diffusional conductance in extracted pore network models using convolutional
  neural networks}.{\BBCQ}
\newblock
\APACjournalVolNumPages{Computers \& Geosciences}{162}{}{105086}.
\PrintBackRefs{\CurrentBib}

\bibitem [\protect \citeauthoryear {%
Nabipour%
, Raoof%
, Cnudde%
, Aghaei%
\BCBL {}\ \BBA {} Qajar%
}{%
Nabipour%
\ \protect \BOthers {.}}{%
{\protect \APACyear {2024}}%
}]{%
nabipour2024computationally}
\APACinsertmetastar {%
nabipour2024computationally}%
\begin{APACrefauthors}%
Nabipour, I.%
, Raoof, A.%
, Cnudde, V.%
, Aghaei, H.%
\BCBL {}\ \BBA {} Qajar, J.%
\end{APACrefauthors}%
\unskip\
\newblock
\APACrefYearMonthDay{2024}{}{}.
\newblock
{\BBOQ}\APACrefatitle {A computationally efficient modeling of flow in complex
  porous media by coupling multiscale digital rock physics and deep learning:
  Improving the tradeoff between resolution and field-of-view} {A
  computationally efficient modeling of flow in complex porous media by
  coupling multiscale digital rock physics and deep learning: Improving the
  tradeoff between resolution and field-of-view}.{\BBCQ}
\newblock
\APACjournalVolNumPages{Advances in Water Resources}{188}{}{104695}.
\PrintBackRefs{\CurrentBib}

\bibitem [\protect \citeauthoryear {%
Prodanovi{\'c}%
\ \protect \BOthers {.}}{%
Prodanovi{\'c}%
\ \protect \BOthers {.}}{%
{\protect \APACyear {2023}}%
}]{%
prodanovic2023digital}
\APACinsertmetastar {%
prodanovic2023digital}%
\begin{APACrefauthors}%
Prodanovi{\'c}, M.%
, Esteva, M.%
, McClure, J.%
, Chang, B\BPBI C.%
, Santos, J\BPBI E.%
, Radhakrishnan, A.%
\BDBL {}Khan, H.%
\end{APACrefauthors}%
\unskip\
\newblock
\APACrefYearMonthDay{2023}{}{}.
\newblock
{\BBOQ}\APACrefatitle {Digital rocks portal (Digital Porous Media): connecting
  data, simulation and community} {Digital rocks portal (digital porous media):
  connecting data, simulation and community}.{\BBCQ}
\newblock
\BIn{} \APACrefbtitle {E3S Web of Conferences} {E3s web of conferences}\
  (\BVOL~367, \BPG~01010).
\PrintBackRefs{\CurrentBib}

\bibitem [\protect \citeauthoryear {%
Raoof%
\ \BBA {} Hassanizadeh%
}{%
Raoof%
\ \BBA {} Hassanizadeh%
}{%
{\protect \APACyear {2010}}%
}]{%
raoof2010new}
\APACinsertmetastar {%
raoof2010new}%
\begin{APACrefauthors}%
Raoof, A.%
\BCBT {}\ \BBA {} Hassanizadeh, S\BPBI M.%
\end{APACrefauthors}%
\unskip\
\newblock
\APACrefYearMonthDay{2010}{}{}.
\newblock
{\BBOQ}\APACrefatitle {A new method for generating pore-network models of
  porous media} {A new method for generating pore-network models of porous
  media}.{\BBCQ}
\newblock
\APACjournalVolNumPages{Transport in porous media}{81}{}{391--407}.
\PrintBackRefs{\CurrentBib}

\bibitem [\protect \citeauthoryear {%
Sakhaee-Pour%
\ \BBA {} Bryant%
}{%
Sakhaee-Pour%
\ \BBA {} Bryant%
}{%
{\protect \APACyear {2012}}%
}]{%
sakhaee2012gas}
\APACinsertmetastar {%
sakhaee2012gas}%
\begin{APACrefauthors}%
Sakhaee-Pour, A.%
\BCBT {}\ \BBA {} Bryant, S\BPBI L.%
\end{APACrefauthors}%
\unskip\
\newblock
\APACrefYearMonthDay{2012}{}{}.
\newblock
{\BBOQ}\APACrefatitle {Gas permeability of shale} {Gas permeability of
  shale}.{\BBCQ}
\newblock
\APACjournalVolNumPages{SPE Reservoir Evaluation \&
  Engineering}{15}{04}{401--409}.
\PrintBackRefs{\CurrentBib}

\bibitem [\protect \citeauthoryear {%
Santos%
\ \protect \BOthers {.}}{%
Santos%
\ \protect \BOthers {.}}{%
{\protect \APACyear {2022}}%
}]{%
santos2022mplbm}
\APACinsertmetastar {%
santos2022mplbm}%
\begin{APACrefauthors}%
Santos, J\BPBI E.%
, Gigliotti, A.%
, Bihani, A.%
, Landry, C.%
, Hesse, M\BPBI A.%
, Pyrcz, M\BPBI J.%
\BCBL {}\ \BBA {} Prodanovi{\'c}, M.%
\end{APACrefauthors}%
\unskip\
\newblock
\APACrefYearMonthDay{2022}{}{}.
\newblock
{\BBOQ}\APACrefatitle {{MPLBM-UT}: Multiphase {LBM} library for permeable media
  analysis} {{MPLBM-UT}: Multiphase {LBM} library for permeable media
  analysis}.{\BBCQ}
\newblock
\APACjournalVolNumPages{SoftwareX}{18}{}{101097}.
\PrintBackRefs{\CurrentBib}

\bibitem [\protect \citeauthoryear {%
Santos%
\ \protect \BOthers {.}}{%
Santos%
\ \protect \BOthers {.}}{%
{\protect \APACyear {2020}}%
}]{%
santos2020poreflow}
\APACinsertmetastar {%
santos2020poreflow}%
\begin{APACrefauthors}%
Santos, J\BPBI E.%
, Xu, D.%
, Jo, H.%
, Landry, C\BPBI J.%
, Prodanovi{\'c}, M.%
\BCBL {}\ \BBA {} Pyrcz, M\BPBI J.%
\end{APACrefauthors}%
\unskip\
\newblock
\APACrefYearMonthDay{2020}{}{}.
\newblock
{\BBOQ}\APACrefatitle {{PoreFlow-Net}: A {3D} convolutional neural network to
  predict fluid flow through porous media} {{PoreFlow-Net}: A {3D}
  convolutional neural network to predict fluid flow through porous
  media}.{\BBCQ}
\newblock
\APACjournalVolNumPages{Advances in Water Resources}{138}{}{103539}.
\PrintBackRefs{\CurrentBib}

\bibitem [\protect \citeauthoryear {%
Scarselli%
, Gori%
, Tsoi%
, Hagenbuchner%
\BCBL {}\ \BBA {} Monfardini%
}{%
Scarselli%
\ \protect \BOthers {.}}{%
{\protect \APACyear {2008}}%
}]{%
scarselli2008graph}
\APACinsertmetastar {%
scarselli2008graph}%
\begin{APACrefauthors}%
Scarselli, F.%
, Gori, M.%
, Tsoi, A\BPBI C.%
, Hagenbuchner, M.%
\BCBL {}\ \BBA {} Monfardini, G.%
\end{APACrefauthors}%
\unskip\
\newblock
\APACrefYearMonthDay{2008}{}{}.
\newblock
{\BBOQ}\APACrefatitle {The graph neural network model} {The graph neural
  network model}.{\BBCQ}
\newblock
\APACjournalVolNumPages{IEEE Transactions on Neural Networks}{20}{1}{61--80}.
\PrintBackRefs{\CurrentBib}

\bibitem [\protect \citeauthoryear {%
Smith%
, Brusseau%
\BCBL {}\ \BBA {} Guo%
}{%
Smith%
\ \protect \BOthers {.}}{%
{\protect \APACyear {2024}}%
}]{%
smith2024integrated}
\APACinsertmetastar {%
smith2024integrated}%
\begin{APACrefauthors}%
Smith, J.%
, Brusseau, M\BPBI L.%
\BCBL {}\ \BBA {} Guo, B.%
\end{APACrefauthors}%
\unskip\
\newblock
\APACrefYearMonthDay{2024}{}{}.
\newblock
{\BBOQ}\APACrefatitle {An integrated analytical modeling framework for
  determining site-specific soil screening levels for {PFAS}} {An integrated
  analytical modeling framework for determining site-specific soil screening
  levels for {PFAS}}.{\BBCQ}
\newblock
\APACjournalVolNumPages{Water Research}{252}{}{121236}.
\PrintBackRefs{\CurrentBib}

\bibitem [\protect \citeauthoryear {%
Tang%
, Zhang%
\BCBL {}\ \BBA {} Li%
}{%
Tang%
\ \protect \BOthers {.}}{%
{\protect \APACyear {2022}}%
}]{%
tang2022predicting}
\APACinsertmetastar {%
tang2022predicting}%
\begin{APACrefauthors}%
Tang, P.%
, Zhang, D.%
\BCBL {}\ \BBA {} Li, H.%
\end{APACrefauthors}%
\unskip\
\newblock
\APACrefYearMonthDay{2022}{}{}.
\newblock
{\BBOQ}\APACrefatitle {Predicting permeability from 3D rock images based on CNN
  with physical information} {Predicting permeability from 3d rock images based
  on cnn with physical information}.{\BBCQ}
\newblock
\APACjournalVolNumPages{Journal of Hydrology}{606}{}{127473}.
\PrintBackRefs{\CurrentBib}

\bibitem [\protect \citeauthoryear {%
Tembely%
, AlSumaiti%
\BCBL {}\ \BBA {} Alameri%
}{%
Tembely%
\ \protect \BOthers {.}}{%
{\protect \APACyear {2020}}%
}]{%
tembely2020deep}
\APACinsertmetastar {%
tembely2020deep}%
\begin{APACrefauthors}%
Tembely, M.%
, AlSumaiti, A\BPBI M.%
\BCBL {}\ \BBA {} Alameri, W.%
\end{APACrefauthors}%
\unskip\
\newblock
\APACrefYearMonthDay{2020}{}{}.
\newblock
{\BBOQ}\APACrefatitle {A deep learning perspective on predicting permeability
  in porous media from network modeling to direct simulation} {A deep learning
  perspective on predicting permeability in porous media from network modeling
  to direct simulation}.{\BBCQ}
\newblock
\APACjournalVolNumPages{Computational Geosciences}{24}{4}{1541--1556}.
\PrintBackRefs{\CurrentBib}

\bibitem [\protect \citeauthoryear {%
Tembely%
, AlSumaiti%
\BCBL {}\ \BBA {} Alameri%
}{%
Tembely%
\ \protect \BOthers {.}}{%
{\protect \APACyear {2021}}%
}]{%
tembely2021machine}
\APACinsertmetastar {%
tembely2021machine}%
\begin{APACrefauthors}%
Tembely, M.%
, AlSumaiti, A\BPBI M.%
\BCBL {}\ \BBA {} Alameri, W\BPBI S.%
\end{APACrefauthors}%
\unskip\
\newblock
\APACrefYearMonthDay{2021}{}{}.
\newblock
{\BBOQ}\APACrefatitle {Machine and deep learning for estimating the
  permeability of complex carbonate rock from {X}-ray micro-computed
  tomography} {Machine and deep learning for estimating the permeability of
  complex carbonate rock from {X}-ray micro-computed tomography}.{\BBCQ}
\newblock
\APACjournalVolNumPages{Energy Reports}{7}{}{1460--1472}.
\PrintBackRefs{\CurrentBib}

\bibitem [\protect \citeauthoryear {%
Tian%
, Qi%
, Sun%
\BCBL {}\ \BBA {} Yaseen%
}{%
Tian%
\ \protect \BOthers {.}}{%
{\protect \APACyear {2020}}%
}]{%
tian2020surrogate}
\APACinsertmetastar {%
tian2020surrogate}%
\begin{APACrefauthors}%
Tian, J.%
, Qi, C.%
, Sun, Y.%
\BCBL {}\ \BBA {} Yaseen, Z\BPBI M.%
\end{APACrefauthors}%
\unskip\
\newblock
\APACrefYearMonthDay{2020}{}{}.
\newblock
{\BBOQ}\APACrefatitle {Surrogate permeability modelling of low-permeable rocks
  using convolutional neural networks} {Surrogate permeability modelling of
  low-permeable rocks using convolutional neural networks}.{\BBCQ}
\newblock
\APACjournalVolNumPages{Computer Methods in Applied Mechanics and
  Engineering}{366}{}{113103}.
\PrintBackRefs{\CurrentBib}

\bibitem [\protect \citeauthoryear {%
Tian%
, Qi%
, Sun%
, Yaseen%
\BCBL {}\ \BBA {} Pham%
}{%
Tian%
\ \protect \BOthers {.}}{%
{\protect \APACyear {2021}}%
}]{%
tian2021permeability}
\APACinsertmetastar {%
tian2021permeability}%
\begin{APACrefauthors}%
Tian, J.%
, Qi, C.%
, Sun, Y.%
, Yaseen, Z\BPBI M.%
\BCBL {}\ \BBA {} Pham, B\BPBI T.%
\end{APACrefauthors}%
\unskip\
\newblock
\APACrefYearMonthDay{2021}{}{}.
\newblock
{\BBOQ}\APACrefatitle {Permeability prediction of porous media using a
  combination of computational fluid dynamics and hybrid machine learning
  methods} {Permeability prediction of porous media using a combination of
  computational fluid dynamics and hybrid machine learning methods}.{\BBCQ}
\newblock
\APACjournalVolNumPages{Engineering with Computers}{37}{}{3455--3471}.
\PrintBackRefs{\CurrentBib}

\bibitem [\protect \citeauthoryear {%
H.~Wang%
\ \protect \BOthers {.}}{%
H.~Wang%
\ \protect \BOthers {.}}{%
{\protect \APACyear {2023}}%
}]{%
wang2023application}
\APACinsertmetastar {%
wang2023application}%
\begin{APACrefauthors}%
Wang, H.%
, Dalton, L.%
, Guo, R.%
, McClure, J.%
, Crandall, D.%
\BCBL {}\ \BBA {} Chen, C.%
\end{APACrefauthors}%
\unskip\
\newblock
\APACrefYearMonthDay{2023}{}{}.
\newblock
{\BBOQ}\APACrefatitle {Application of unsupervised deep learning to image
  segmentation and in-situ contact angle measurements in a CO2-water-rock
  system} {Application of unsupervised deep learning to image segmentation and
  in-situ contact angle measurements in a co2-water-rock system}.{\BBCQ}
\newblock
\APACjournalVolNumPages{Advances in Water Resources}{173}{}{104385}.
\PrintBackRefs{\CurrentBib}

\bibitem [\protect \citeauthoryear {%
Y.~Wang%
, Chakrapani%
, Wen%
\BCBL {}\ \BBA {} Hajibeygi%
}{%
Y.~Wang%
\ \protect \BOthers {.}}{%
{\protect \APACyear {2024}}%
}]{%
wang2024pore}
\APACinsertmetastar {%
wang2024pore}%
\begin{APACrefauthors}%
Wang, Y.%
, Chakrapani, T\BPBI H.%
, Wen, Z.%
\BCBL {}\ \BBA {} Hajibeygi, H.%
\end{APACrefauthors}%
\unskip\
\newblock
\APACrefYearMonthDay{2024}{}{}.
\newblock
{\BBOQ}\APACrefatitle {Pore-scale simulation of H2-brine system relevant for
  underground hydrogen storage: A lattice Boltzmann investigation} {Pore-scale
  simulation of h2-brine system relevant for underground hydrogen storage: A
  lattice boltzmann investigation}.{\BBCQ}
\newblock
\APACjournalVolNumPages{Advances in Water Resources}{190}{}{104756}.
\PrintBackRefs{\CurrentBib}

\bibitem [\protect \citeauthoryear {%
Y\BPBI D.~Wang%
, Blunt%
, Armstrong%
\BCBL {}\ \BBA {} Mostaghimi%
}{%
Y\BPBI D.~Wang%
, Blunt%
\BCBL {}\ \protect \BOthers {.}}{%
{\protect \APACyear {2021}}%
}]{%
da2021deep}
\APACinsertmetastar {%
da2021deep}%
\begin{APACrefauthors}%
Wang, Y\BPBI D.%
, Blunt, M\BPBI J.%
, Armstrong, R\BPBI T.%
\BCBL {}\ \BBA {} Mostaghimi, P.%
\end{APACrefauthors}%
\unskip\
\newblock
\APACrefYearMonthDay{2021}{}{}.
\newblock
{\BBOQ}\APACrefatitle {Deep learning in pore scale imaging and modeling} {Deep
  learning in pore scale imaging and modeling}.{\BBCQ}
\newblock
\APACjournalVolNumPages{Earth-Science Reviews}{215}{}{103555}.
\PrintBackRefs{\CurrentBib}

\bibitem [\protect \citeauthoryear {%
Y\BPBI D.~Wang%
, Chung%
, Armstrong%
\BCBL {}\ \BBA {} Mostaghimi%
}{%
Y\BPBI D.~Wang%
, Chung%
\BCBL {}\ \protect \BOthers {.}}{%
{\protect \APACyear {2021}}%
}]{%
wang2021ml}
\APACinsertmetastar {%
wang2021ml}%
\begin{APACrefauthors}%
Wang, Y\BPBI D.%
, Chung, T.%
, Armstrong, R\BPBI T.%
\BCBL {}\ \BBA {} Mostaghimi, P.%
\end{APACrefauthors}%
\unskip\
\newblock
\APACrefYearMonthDay{2021}{}{}.
\newblock
{\BBOQ}\APACrefatitle {{ML-LBM}: predicting and accelerating steady state flow
  simulation in porous media with convolutional neural networks} {{ML-LBM}:
  predicting and accelerating steady state flow simulation in porous media with
  convolutional neural networks}.{\BBCQ}
\newblock
\APACjournalVolNumPages{Transport in Porous Media}{138}{1}{49--75}.
\PrintBackRefs{\CurrentBib}

\bibitem [\protect \citeauthoryear {%
J.~Wu%
, Yin%
\BCBL {}\ \BBA {} Xiao%
}{%
J.~Wu%
\ \protect \BOthers {.}}{%
{\protect \APACyear {2018}}%
}]{%
wu2018seeing}
\APACinsertmetastar {%
wu2018seeing}%
\begin{APACrefauthors}%
Wu, J.%
, Yin, X.%
\BCBL {}\ \BBA {} Xiao, H.%
\end{APACrefauthors}%
\unskip\
\newblock
\APACrefYearMonthDay{2018}{}{}.
\newblock
{\BBOQ}\APACrefatitle {Seeing permeability from images: fast prediction with
  convolutional neural networks} {Seeing permeability from images: fast
  prediction with convolutional neural networks}.{\BBCQ}
\newblock
\APACjournalVolNumPages{Science Bulletin}{63}{18}{1215--1222}.
\PrintBackRefs{\CurrentBib}

\bibitem [\protect \citeauthoryear {%
Z.~Wu%
\ \protect \BOthers {.}}{%
Z.~Wu%
\ \protect \BOthers {.}}{%
{\protect \APACyear {2020}}%
}]{%
wu2020comprehensive}
\APACinsertmetastar {%
wu2020comprehensive}%
\begin{APACrefauthors}%
Wu, Z.%
, Pan, S.%
, Chen, F.%
, Long, G.%
, Zhang, C.%
\BCBL {}\ \BBA {} Yu, P\BPBI S.%
\end{APACrefauthors}%
\unskip\
\newblock
\APACrefYearMonthDay{2020}{}{}.
\newblock
{\BBOQ}\APACrefatitle {A comprehensive survey on graph neural networks} {A
  comprehensive survey on graph neural networks}.{\BBCQ}
\newblock
\APACjournalVolNumPages{IEEE Transactions on Neural Networks and Learning
  Systems}{32}{1}{4--24}.
\PrintBackRefs{\CurrentBib}

\bibitem [\protect \citeauthoryear {%
Yang%
\ \protect \BOthers {.}}{%
Yang%
\ \protect \BOthers {.}}{%
{\protect \APACyear {2024}}%
}]{%
yang2024data}
\APACinsertmetastar {%
yang2024data}%
\begin{APACrefauthors}%
Yang, G.%
, Xu, R.%
, Tian, Y.%
, Guo, S.%
, Wu, J.%
\BCBL {}\ \BBA {} Chu, X.%
\end{APACrefauthors}%
\unskip\
\newblock
\APACrefYearMonthDay{2024}{}{}.
\newblock
{\BBOQ}\APACrefatitle {Data-driven methods for flow and transport in porous
  media: A review} {Data-driven methods for flow and transport in porous media:
  A review}.{\BBCQ}
\newblock
\APACjournalVolNumPages{International Journal of Heat and Mass
  Transfer}{235}{}{126149}.
\PrintBackRefs{\CurrentBib}

\bibitem [\protect \citeauthoryear {%
H.~Zhang%
\ \protect \BOthers {.}}{%
H.~Zhang%
\ \protect \BOthers {.}}{%
{\protect \APACyear {2022}}%
}]{%
zhang2022permeability}
\APACinsertmetastar {%
zhang2022permeability}%
\begin{APACrefauthors}%
Zhang, H.%
, Yu, H.%
, Yuan, X.%
, Xu, H.%
, Micheal, M.%
, Zhang, J.%
\BDBL {}Wu, H.%
\end{APACrefauthors}%
\unskip\
\newblock
\APACrefYearMonthDay{2022}{}{}.
\newblock
{\BBOQ}\APACrefatitle {Permeability prediction of low-resolution porous media
  images using autoencoder-based convolutional neural network} {Permeability
  prediction of low-resolution porous media images using autoencoder-based
  convolutional neural network}.{\BBCQ}
\newblock
\APACjournalVolNumPages{Journal of Petroleum Science and
  Engineering}{208}{}{109589}.
\PrintBackRefs{\CurrentBib}

\bibitem [\protect \citeauthoryear {%
M.~Zhang%
\ \BBA {} Chen%
}{%
M.~Zhang%
\ \BBA {} Chen%
}{%
{\protect \APACyear {2018}}%
}]{%
zhang2018link}
\APACinsertmetastar {%
zhang2018link}%
\begin{APACrefauthors}%
Zhang, M.%
\BCBT {}\ \BBA {} Chen, Y.%
\end{APACrefauthors}%
\unskip\
\newblock
\APACrefYearMonthDay{2018}{}{}.
\newblock
{\BBOQ}\APACrefatitle {Link prediction based on graph neural networks} {Link
  prediction based on graph neural networks}.{\BBCQ}
\newblock
\BIn{} \APACrefbtitle {Advances in Neural Information Processing Systems}
  {Advances in neural information processing systems}\ (\BVOL~31, \BPGS\
  5165--5175).
\PrintBackRefs{\CurrentBib}

\bibitem [\protect \citeauthoryear {%
X\BHBI L.~Zhang%
, Xiao%
, Luo%
\BCBL {}\ \BBA {} He%
}{%
X\BHBI L.~Zhang%
\ \protect \BOthers {.}}{%
{\protect \APACyear {2022}}%
}]{%
zhang2022ensemble}
\APACinsertmetastar {%
zhang2022ensemble}%
\begin{APACrefauthors}%
Zhang, X\BHBI L.%
, Xiao, H.%
, Luo, X.%
\BCBL {}\ \BBA {} He, G.%
\end{APACrefauthors}%
\unskip\
\newblock
\APACrefYearMonthDay{2022}{}{}.
\newblock
{\BBOQ}\APACrefatitle {Ensemble {Kalman} method for learning turbulence models
  from indirect observation data} {Ensemble {Kalman} method for learning
  turbulence models from indirect observation data}.{\BBCQ}
\newblock
\APACjournalVolNumPages{Journal of Fluid Mechanics}{949}{}{A26}.
\PrintBackRefs{\CurrentBib}

\bibitem [\protect \citeauthoryear {%
Q.~Zhao%
, Han%
, Guo%
\BCBL {}\ \BBA {} Chen%
}{%
Q.~Zhao%
\ \protect \BOthers {.}}{%
{\protect \APACyear {2025}}%
}]{%
zhao2025computationally}
\APACinsertmetastar {%
zhao2025computationally}%
\begin{APACrefauthors}%
Zhao, Q.%
, Han, X.%
, Guo, R.%
\BCBL {}\ \BBA {} Chen, C.%
\end{APACrefauthors}%
\unskip\
\newblock
\APACrefYearMonthDay{2025}{}{}.
\newblock
{\BBOQ}\APACrefatitle {A computationally efficient hybrid neural network
  architecture for porous media: Integrating convolutional and graph neural
  networks for improved property predictions} {A computationally efficient
  hybrid neural network architecture for porous media: Integrating
  convolutional and graph neural networks for improved property
  predictions}.{\BBCQ}
\newblock
\APACjournalVolNumPages{Advances in Water Resources}{195}{}{104881}.
\PrintBackRefs{\CurrentBib}

\bibitem [\protect \citeauthoryear {%
X.~Zhao%
, Zhong%
\BCBL {}\ \BBA {} Li%
}{%
X.~Zhao%
\ \protect \BOthers {.}}{%
{\protect \APACyear {2024}}%
}]{%
zhao2024rtg}
\APACinsertmetastar {%
zhao2024rtg}%
\begin{APACrefauthors}%
Zhao, X.%
, Zhong, Y.%
\BCBL {}\ \BBA {} Li, P.%
\end{APACrefauthors}%
\unskip\
\newblock
\APACrefYearMonthDay{2024}{}{}.
\newblock
{\BBOQ}\APACrefatitle {{RTG-GNN}: A novel rock topology-guided approach for
  permeability prediction using graph neural networks} {{RTG-GNN}: A novel rock
  topology-guided approach for permeability prediction using graph neural
  networks}.{\BBCQ}
\newblock
\APACjournalVolNumPages{Geoenergy Science and Engineering}{243}{}{213358}.
\PrintBackRefs{\CurrentBib}

\bibitem [\protect \citeauthoryear {%
Zhou%
, McClure%
, Chen%
\BCBL {}\ \BBA {} Xiao%
}{%
Zhou%
\ \protect \BOthers {.}}{%
{\protect \APACyear {2022}}%
}]{%
zhou2022neural}
\APACinsertmetastar {%
zhou2022neural}%
\begin{APACrefauthors}%
Zhou, X\BHBI H.%
, McClure, J\BPBI E.%
, Chen, C.%
\BCBL {}\ \BBA {} Xiao, H.%
\end{APACrefauthors}%
\unskip\
\newblock
\APACrefYearMonthDay{2022}{}{}.
\newblock
{\BBOQ}\APACrefatitle {Neural network--based pore flow field prediction in
  porous media using super resolution} {Neural network--based pore flow field
  prediction in porous media using super resolution}.{\BBCQ}
\newblock
\APACjournalVolNumPages{Physical Review Fluids}{7}{7}{074302}.
\PrintBackRefs{\CurrentBib}

\end{thebibliography}

%Reference citation instructions and examples:
%
% Please use ONLY \cite and \citeA for reference citations.
% \cite for parenthetical references
% ...as shown in recent studies (Simpson et al., 2019)
% \citeA for in-text citations
% ...Simpson et al. (2019) have shown...
%
%
%...as shown by \citeA{jskilby}.
%...as shown by \citeA{lewin76}, \citeA{carson86}, \citeA{bartoldy02}, and \citeA{rinaldi03}.
%...has been shown \cite{jskilbye}.
%...has been shown \cite{lewin76,carson86,bartoldy02,rinaldi03}.
%... \cite <i.e.>[]{lewin76,carson86,bartoldy02,rinaldi03}.
%...has been shown by \cite <e.g.,>[and others]{lewin76}.
%
% apacite uses < > for prenotes and [ ] for postnotes
% DO NOT use other cite commands (e.g., \citet, \citep, \citeyear, \nocite, \citealp, etc.).
%

\end{document}